%% file: HRI2021.tex
\documentclass[sigconf]{acmart}

\AtBeginDocument{%
  \providecommand\BibTeX{{%
    \normalfont B\kern-0.5em{\scshape i\kern-0.25em b}\kern-0.8em\TeX}}}

\copyrightyear{2021}
\acmYear{2021}
\setcopyright{rightsretained}
\acmConference[HRI '21]{Proceedings of the 2021 ACM/IEEE International Conference on Human-Robot Interaction}{March 8--11, 2021}{Boulder, CO, USA}
\acmBooktitle{Proceedings of the 2021 ACM/IEEE International Conference on Human-Robot Interaction (HRI '21), March 8--11, 2021, Boulder, CO, USA}
\acmDOI{10.1145/3434073.3444667}
\acmISBN{978-1-4503-8289-2/21/03}

\usepackage[utf8]{inputenc} 
\usepackage[T1]{fontenc}    
\usepackage{hyperref}       
\usepackage{url}            
\usepackage{booktabs}       
\usepackage{amsfonts}       
\usepackage{amsmath}
\usepackage{nicefrac}       
\usepackage{microtype}      
\usepackage[ruled,noend]{algorithm2e}
\usepackage{xcolor}
\usepackage[nolist,nohyperlinks]{acronym}
\SetArgSty{textnormal}
\usepackage{graphicx}
\usepackage{caption}
\usepackage{subcaption}
\usepackage[export]{adjustbox}
\usepackage{float}
\usepackage{balance}

\input{macros.tex}

\begin{acronym}
\acro{IRL}[IRL]{Inverse Reinforcement Learning}
\acro{FERL}{Feature Expansive Reward Learning}
\acro{ME-IRL}{Maximum Entropy Inverse Reinforcement Learning}
\acro{MDP}{Markov Decision Process}
\acro{POMDP}{Partially Observable Markov Decision Process}
\acro{MAP}{Maximum A Posterior}
\acro{GAN}{Generative Adversarial Networks}
\acro{EE}{End-Effector}
\acro{MSE}{Mean-Squared-Error}
\acro{GT}{Ground Truth}
\end{acronym}
\usepackage{soul}

\newcommand{\abnote}[1]{\textcolor{blue}{AB: #1}}

\newcommand{\change}[1]{\textcolor{black}{#1}}

\begin{document}
\fancyhead{}

\title{Feature Expansive Reward Learning: \\Rethinking Human Input}


\author{Andreea Bobu}
\authornote{
    Both authors contributed equally to this research.\\
    This research is supported by the Air Force Office of Scientific Research (AFOSR), the Office of Naval Research (ONR-YIP), the DARPA Assured Autonomy Grant, the CONIX Research Center, and the German Academic Exchange Service (DAAD).
}
\affiliation{%
  \institution{University of California, Berkeley}
}
\email{abobu@berkeley.edu}

\author{Marius Wiggert}
\authornotemark[1]
\affiliation{%
  \institution{University of California, Berkeley}
}
\email{mariuswiggert@berkeley.edu}

\author{Claire Tomlin}
\affiliation{%
  \institution{University of California, Berkeley}
}
\email{tomlin@berkeley.edu}

\author{Anca D. Dragan}
\affiliation{%
 \institution{University of California, Berkeley}
}
\email{anca@berkeley.edu}

\renewcommand{\shortauthors}{Bobu and Wiggert, et al.}

\begin{abstract}
When a person is not satisfied with how a robot performs a task, they can intervene to correct it. 
Reward learning methods enable the robot to adapt its reward function online based on such human input, but they rely on handcrafted features. 
When the correction cannot be explained by these features,
recent work in deep \ac{IRL} suggests that the robot could ask for task demonstrations and recover a reward defined over the raw state space. Our insight is that rather than \textit{implicitly} learning about the missing feature(s) from demonstrations, the robot should instead ask for data that \textit{explicitly} teaches it about what it is missing. 
We introduce a new type of human input in which the person guides the robot from states where the feature being taught is highly expressed to states where it is not. We propose an algorithm for learning the feature from the raw state space and integrating it into the reward function.
By focusing the human input on the missing feature, our method decreases sample complexity and improves generalization of the learned reward over the above deep \ac{IRL} baseline. We show this in experiments with a physical 7DOF robot manipulator, as well as in a user study conducted in a simulated environment. 
\end{abstract}


\maketitle

\input{1_intro}
\input{2_method}

\input{3_FERL_experts}
\input{4_FERL_users}

\input{5_discussion}
\bibliographystyle{ACM-Reference-Format}
\bibliography{HRI2021}


\newpage

\appendix

\input{6_supplement}

\end{document}

%% file: macros.tex
\newcommand{\sysstate}{\ensuremath{s}}
\newcommand{\action}{a}

\newcommand{\trace}{\ensuremath{\xi}}
\newcommand{\traj}{\ensuremath{\tau}}

%% file: 1_intro.tex
\begin{figure}
\includegraphics[width=0.47\textwidth]{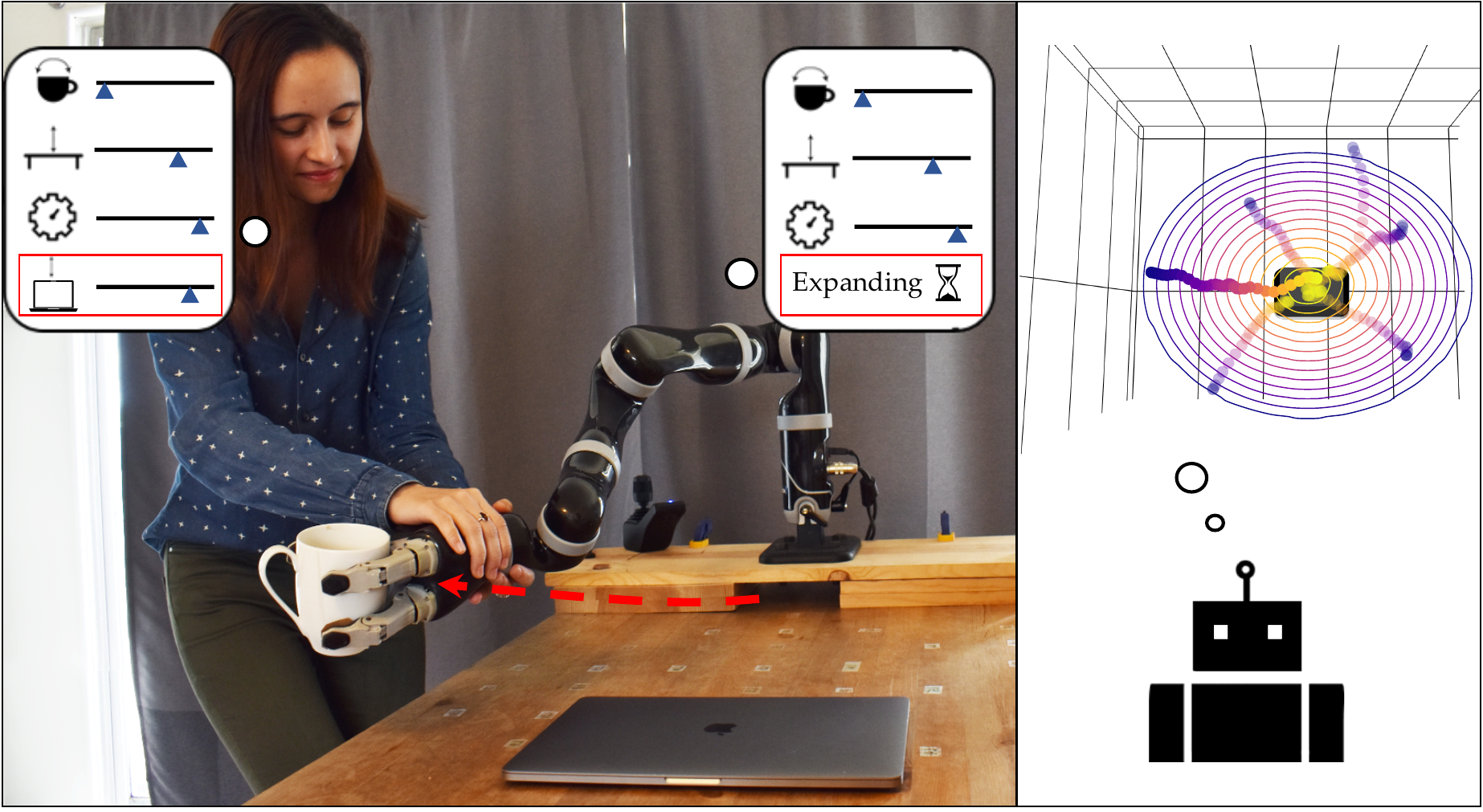}
\centering
\caption{(Left) After the robot detects that its feature space cannot explain the human's input, the person attempts to teach it the concept of distance from the laptop. (Right) The robot queries the human for feature traces that teach it the missing feature and adapts the reward to account for it.}
\label{fig:front_fig}
\end{figure}

\section{Introduction}


When we deploy robots in human environments, they have to be able to adapt their reward functions to human preferences. For instance in the scenario in Fig. \ref{fig:front_fig}, the robot was carrying the cup over the laptop, risking a spill, and the person intervened to correct it. Recent methods interpret such corrections as evidence about the human's desired preference for how to complete the task, enabling the robot to update its reward function online \cite{bajcsy2017phri, jain2015learning, bajcsy2018onefeature}.

Because they have to perform updates online from very little input, these methods resort to representing the reward as a linear function of a small set of hand-engineered \textit{features}. Unfortunately, this puts too much burden on system designers: 
specifying \textit{a priori} an exhaustive set of \emph{all} the features that end-users might care about is impossible for real-world tasks. While prior work has enabled robots to at least \emph{detect} that the features it has access to cannot explain the human's input \cite{bobu2020quantifying}, it is still unclear how the robot might then construct a feature that can explain it.
A natural answer is in deep \ac{IRL} methods \cite{wulfmeier2016maxentirl, finn2016gcl, levine2011nonlinear}, which learn rewards defined directly on the high-dimensional raw state (or observation) space, thereby constructing features automatically. When the robot can't make sense of the human input, it can ask for demonstrations of the task, and learn a reward over not just the known features, but also the raw state space. On the bright side, the learned reward would now be able to explain the demonstrations. On the downside, this may come at the cost of losing generalization when venturing sufficiently far from the demonstrations \cite{fu2018learning, Reddy2020SQILIL}.

In this work, we propose an alternative to relying on demonstrations: we can co-design the learner together with the \emph{type of human feedback} we ask for. Our insight is that instead of learning about the missing feature(s) \emph{implicitly} through the optimal actions, the robot should ask for data that \emph{explicitly} teaches it what is missing. We introduce a new type of human input, which we call \emph{feature traces} -- partial trajectories that describe the monotonic evolution of the value of the feature to be learned. To provide a feature trace, the person guides the robot from states where the missing feature is highly expressed to states where it is not, in a monotonic fashion.

Looking back at Fig. \ref{fig:front_fig}, the person teaches the robot to avoid the laptop by giving a few feature traces: she starts with the arm above the laptop and moves it away until comfortable with the distance from the object. We introduce a reward learning algorithm that harvests the structure inherent to feature traces and uses it to efficiently train a generalizable aspect of the reward: in our example, the distance from the laptop. In experiments on a 7DoF robot arm and in a user study, we find that by taking control of not only the algorithm but also the kind of data it can receive, our method is able to recover more generalizable rewards with much less human input compared to a learning from demonstrations baseline.
\change{Although showcased in manipulation, our method should be useful in any scenarios where it is difficult to pre-specify the features the person may care about: in collaborative manufacturing users might care about the rotation of the object handed over, although the robot was only programmed to consider the position, or in autonomous driving passengers may care about how fast to drive through curves.}


Finally, we discuss our method's potential implications for the general deep reward learning community. Feature traces enable humans to teach robots about salient aspects of the reward in an intuitive way, making it easier to learn overall rewards. This suggests that taking a divide-and-conquer approach focusing on learning important features separately before learning the reward could benefit \ac{IRL} generalization and sample complexity. Although more work is needed to teach difficult features with even less supervision, we are excited to have taken a step towards better disambiguating complex reward functions that explain human inputs such as demonstrations in the context of reward learning.

%% file: 2_method.tex
\section{Method}


A standard way to teach the robot a reward function over raw state is to provide demonstrations and have the robot learn via deep IRL \cite{wulfmeier2016maxentirl, finn2016gcl}. Our method introduces a divide-and-conquer alternative, where users can teach the robot explicitly about features that are important, rather than implicitly through full task demonstrations. 
We assume the robot has access to an initial feature set defined over states $\vec\phi(\sysstate)$, and is optimizing its current estimate of the reward function, $r_{\theta}(\vec\phi(\sysstate))$. Here, $\theta$ is a vector of parameters specifying how the features combine. 
If the robot is not executing the task according to the person's preferences, the human can intervene with input $a_H$. For instance, $a_H$ might be an external torque that the person applies to change the robot's current configuration. Or, they might stop the robot and kinesthetically demonstrate the task, resulting in a trajectory. Building on prior work, we assume the robot can evaluate whether its existing feature space can explain the human input (Sec.~\ref{sec:confidenceestimation}). If it can, the robot directly updates its reward function parameters $\theta$, also in line with prior work~\cite{bajcsy2017phri,ratliff2006MMP} (Sec. \ref{sec:rewardupdate}). But what if it can not? Below, we introduce a method for augmenting the robot's feature space by soliciting further feature-specific human input (Sec. \ref{sec:humaninput}) and using it to learn a new mapping directly from the raw state space (Sec. \ref{sec:featurelearning}). \footnote{Prior work proposed tackling this issue by determining which linear feature to add \cite{haug2018teaching}, 
iterative boosting \cite{ratliff2007boosting},
 or constructing binary features \cite{levine2010feature,choi2013bayesian}. We instead focus on unknown features that can be non-binary, non-linear functions of the raw state.}

\subsection{Collecting Feature Traces from Humans}
\label{sec:humaninput}

A state feature is an arbitrary complex mapping $\phi(\sysstate): \mathbb{R}^d \to [0,1]$. To teach a non-linear representation of $\phi$ with little data, we have to balance getting informative inputs and not placing too much burden on the human. As such, we introduce \textit{feature traces} $\trace = \sysstate_{0:n}$, a novel type of human input defined as a sequence of $n$ states monotonically decreasing in feature value, i.e. $\phi(\sysstate_i) \geq \phi(\sysstate_j), \forall i < j$. When learning a feature, the robot queries the human for a set $\Xi$ of $N$ traces. The person gives a trace by simply moving the system from any state $\sysstate_0$ to an end state, noisily ensuring monotonicity. 
\change{Our method, thus, only requires an interface for communicating ordered preferences over states: kinesthetic teaching is useful for household or small industrial robots, while teleoperation and simulation interfaces may be better for larger systems.}
Because feature learning was triggered by a correction, it is fair to assume that the human knows what aspect of the task they were trying to correct. 

To illustrate how a human might offer feature traces, let's turn again to Fig. \ref{fig:front_fig}. Here, the person is teaching the robot to keep the mug away from the laptop. The person starts a trace at $\sysstate_0$ by placing the end-effector close to the object center, then leads the robot away from the laptop to $\sysstate_n$. Our method works best when the person tries to be informative, i.e. covers diverse areas of the space: the traces illustrated move radially in all directions and start at different heights. While for some features, like distance from an object, it is easy to be informative, for others, like slowing down near objects, it might be more difficult. We explore how easy it is for users to be informative in our study in Sec. \ref{sec:FERL_users}, with encouraging findings. Further, this limitation can be potentially alleviated using active learning, thereby shifting the burden away from the human to select informative traces, onto the robot to make queries for traces by proposing informative starting states. For instance, the robot could fit an ensemble of functions from traces online, and query for new traces from states where the ensemble disagrees \cite{reddy2019learning}.

The power of feature traces lies in their inherent structure. 
Our algorithm, thus, makes certain assumptions to harvest this structure for learning. 
First, we assume that the feature values of states along the collected traces $\trace \in \Xi$ are monotonically decreasing. 
Since humans are imperfect, we allow users to violate this assumption by modeling them as noisily rational, following the classic Bradley Terry and Luce-Shepard models of preference \cite{bradley1952rank, Luce1959choice}: 
\begin{equation}
    P(\sysstate \succ \sysstate') = P(\phi(\sysstate) > \phi(\sysstate')) = \frac{e^{\phi(\sysstate)}}{e^{\phi(\sysstate)} + e^{\phi(\sysstate')}} \enspace.
    \label{eq:softmax}
\end{equation}
Our method also assumes by default that $\phi(\sysstate_{0}) \approx 1$ and $\phi(\sysstate_{n})\approx 0$, meaning the human starts in a state $s_0$ where the missing feature is highly expressed, then leads the system to a state $\sysstate_{n}$ along decreasing feature values. Since in some situations providing a 1-0 trace is difficult, our algorithm optionally allows the human to give labels $l_{0}, l_{n} \in [0,1]$\footnote{Since providing decimal labels is difficult, the person gives a rating between 0 and 10.} for the respective feature values.





\subsection{Learning a Feature Function}
\label{sec:featurelearning}

We represent the missing feature by a neural network $\phi_{\psi}(\sysstate) : \mathbb{R}^d \to [0,1]$ . We use the feature traces $\trace$, their inherent monotonicity, and the approximate knowledge about $\phi(\sysstate_{0})$ and $\phi(\sysstate_{n})$ to train $\phi_\psi$.

Using Eq. \ref{eq:softmax}, we frame feature learning as a classification problem with a Maximum Likelihood objective over a dataset of tuples $(\sysstate, \sysstate', y) \in \mathcal{D}$, where $y \in \{0,0.5,1\}$ is a label indicating which state has higher feature value. First, due to monotonicity along a feature trace $\trace = (\sysstate_0, \sysstate_1, \dots, \sysstate_n)$, we have $\phi_{\psi}(\sysstate_i) \geq \phi_{\psi}(\sysstate_j), \forall j>i$, so $y=1$ if $j>i$ and 0 otherwise. This results in $\binom{(n+1)}{2}$ tuples per trace, \change{which encourage feature values to decrease monotonically along a trace, but they alone don't enforce the same start and end values across traces.}
Thus, we encode that $\phi(\sysstate_{0}) \approx 1$ and $\phi(\sysstate_{n}) \approx 0$ for all $\trace \in \Xi$ by encouraging indistinguishable feature values at the starts and ends of traces as $(\sysstate_{0}^i, \sysstate_{0}^j, 0.5), (\sysstate_{n}^i, \sysstate_{n}^j, 0.5) \; \forall \; \trace_i, \trace_j \in \Xi, \;i \neq j,\; i > j$. 
This results in a total dataset of $|\mathcal{D}| = \sum_{i=1}^{N}\binom{(n^i+1)}{2} + 2 \binom{N}{2}$ that is already significantly large for a small set of feature traces. 
The final cross-entropy loss $L(\psi)$ is then given by:
\begin{align}
    - \sum_{(\sysstate, \sysstate',y) \in \mathcal{D}} y\log(P(\sysstate \succ \sysstate')) + (1-y)\log(1-P(\sysstate \succ \sysstate'))
    \label{eq:loss}
\end{align}
which expands into the sum of a monotonicity loss for $\sysstate \succ \sysstate'$ and an indistinguishability loss for $\sysstate \sim \sysstate'$:
\begin{align}
    -\sum_{\sysstate \succ \sysstate'} \log\frac{e^{\phi_{\psi}(\sysstate)}}{e^{\phi_{\psi}(\sysstate')} + e^{\phi_{\psi}(\sysstate)}} 
    - \frac{1}{2}\sum_{\sysstate \sim \sysstate'} \log \frac{e^{\phi_{\psi}(\sysstate)+\phi_{\psi}(\sysstate')}}{(e^{\phi_{\psi}(\sysstate)} + e^{\phi_{\psi}(\sysstate')})^2}.
    \label{eq:loss_tuples}
\end{align}
The optional labels $l_{0},l_{n}$ are incorporated as bonus for $\sysstate_{0}, \sysstate_{n}$ as $\phi_{\psi}(\sysstate_{0})' = \phi_{\psi}(\sysstate_{0}) - l_{0}$ and $\phi_{\psi}(\sysstate_{n})' = \phi_{\psi}(\sysstate_{n}) + (1-l_{n})$ to encourage the labeled feature values to approach the labels.
\change{Similar preference learning has been used in imitation and reward learning~\cite{christiano2017preferences,Ibarz2018reward}. The key differences are that the loss in \eqref{eq:loss_tuples} is over feature functions not rewards, and that preferences are provided via feature traces.}

\subsection{Online Reward Update}
\label{sec:rewardupdate}



Once we have a new feature, the robot updates its feature vector $\vec\phi \gets (\vec\phi, \phi_\psi)$. At this point, the robot goes back to the original human input $a_H$ that previously could not be explained by the old features and uses it to update its estimate of the reward parameters $\theta$. Here, any prior work on online reward learning from user input is applicable, but we highlight one example to complete the exposition.

\begin{algorithm}
\DontPrintSemicolon
\textbf{Input:} Features $\vec\phi=[\phi_1, \dots, \phi_f]$, weight $\theta$, confidence threshold $\epsilon$, robot trajectory $\traj$, $N$ number of queries. \\
\While{executing $\traj$}{
    \If{$\action_H$}{
    $\hat\beta \gets$ estimate\_confidence($\action_H$) as in Sec. \ref{sec:confidenceestimation}.\\
         \If{$\beta < \epsilon$}{
        $\Xi = \{\}$\\
        \For{$i\leftarrow 1$ \KwTo $N$}{
            $\trace \gets$ query\_feature\_trace() as in Sec. \ref{sec:humaninput}.\\
            $\Xi \gets \Xi \cup \trace$.
        }
        $\phi_{new} \gets$ learn\_feature($\Xi$) as in Sec. \ref{sec:featurelearning}.\\
        $\vec{\phi} \gets (\vec{\phi}, \phi_{new}), \theta \gets (\theta, 0.0)$.
         }
          $\theta \gets$ update\_reward($\action_H$) as in Sec. \ref{sec:rewardupdate}.\\
          $\traj \gets$ replan\_trajectory($\theta$).
    }
}
\caption{\ac{FERL}}
\label{alg:FERL_overview}
\end{algorithm}

For instance, take the setting where the human's input $a_H$ was an external torque, applied as the robot was tracking a trajectory $\tau$ that was optimal under its current reward. Prior work \cite{bajcsy2017phri} has modeled this as inducing a deformed trajectory $\tau_H$, by propagating the change in configuration to the rest of the trajectory. Further, let $\theta$ define linear weights on the features. Then, the robot updates its estimate $\hat\theta$ in the direction of the feature change from $\tau$ to $\tau_H$
\begin{equation}
    \hat\theta' = \hat\theta - \alpha \left( \sum_{\sysstate \in \traj_H}\vec\phi(\sysstate) -  \sum_{\sysstate \in \traj}\vec\phi(\sysstate)\right) = \hat\theta - \alpha\Big(\Phi(\traj_H) - \Phi(\traj)\Big) \enspace ,
    \label{eq:thetaupdate}
\end{equation}
where $\Phi$ is the cumulative feature sum along a trajectory. If instead, the human intervened with a full demonstration, work on online learning from demonstrations (Sec. 3.2 in \cite{ratliff2006MMP}) has derived the same update with $\tau_H$ now the human demonstration. In our implementation, we use corrections and follow \cite{bajcsy2018onefeature}, which shows that people more easily correct one feature at a time, and only update the $\theta$ index corresponding to the feature that changes the most (after feature learning this is the newly learned feature). After the update, the robot replans its trajectory using the new reward.

\subsection{Confidence Estimation} 
\label{sec:confidenceestimation}

We lastly have to detect that a feature is missing in the first place. Prior work does so by looking at how people's choices are modeled via the Boltzmann noisily-rational decision model:
\begin{equation}
    P(\traj_H \mid \theta, \beta) = \frac{e^{\beta R_{\theta}(\traj_H)}}{\int_{\bar{\traj}_H}e^{\beta R_{\theta}(\bar{\traj}_H)} d\bar{\traj}_H} \enspace ,
\end{equation}
where the human picks trajectories proportional to their exponentiated reward \cite{baker2007goal,von1945theory}.
Here, $\beta \in [0, \infty)$ controls how much the robot expects to observe human input consistent with its feature space. Typically, $\beta$ is fixed, recovering the Maximum Entropy \ac{IRL} \cite{maxent} observation model. Inspired by work in \cite{fridovich-keil2019confidence,fisac2018probabilistically,bobu2020quantifying}, we instead reinterpret it as a confidence in the robot's features' ability to explain human data. To detect missing features, we estimate $\hat\beta$ via a Bayesian belief update $b'(\theta, \beta) \propto P(\traj_H \mid \theta, \beta)b(\theta, \beta)$. If $\hat\beta$ is above a threshold $\epsilon$, the robot updates the reward as usual with its current features; if $\hat\beta < \epsilon$, the features are insufficient and the robot enters feature learning mode. Algorithm \ref{alg:FERL_overview} summarizes the full procedure.

%% file: 3_FERL_experts.tex
\section{FERL Analysis with Expert Users}
\label{sec:FERL_experts}

We first analyze our method (\ac{FERL}) with real robot data collected from an expert -- \change{a person familiar with how the algorithm works} -- to sanity check its benefits relative to standard reward learning. We will test \ac{FERL} with \change{non-experts -- people not familiar with \ac{FERL} but are taught to use it --} in a user study in Sec. \ref{sec:FERL_users}.

\subsection{Feature Learning}
\label{sec:feature_expert}

Before investigating the benefits of \ac{FERL} for reward learning, we analyze the quality of the features it can learn. We present results for six different features of varying complexity.

\begin{figure}
\centering
\begin{subfigure}{.19\textwidth}
  \centering
  \includegraphics[width=\textwidth,left]{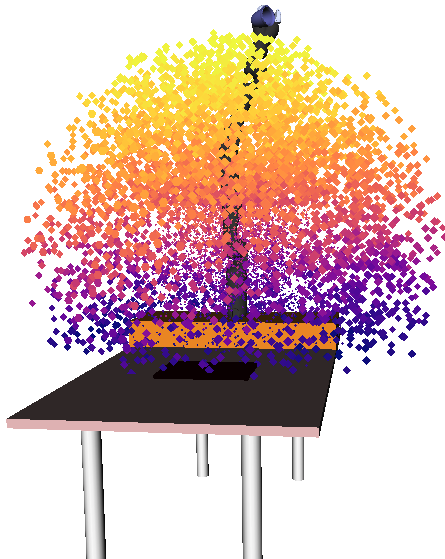}
\end{subfigure}%
\begin{subfigure}{.27\textwidth}
  \centering
  \includegraphics[width=\textwidth,left]{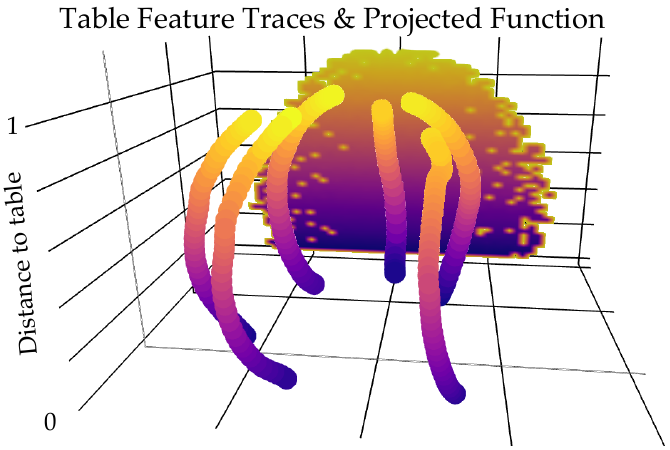}
\end{subfigure}
\begin{subfigure}{.19\textwidth}
  \centering
  \includegraphics[width=\textwidth,left]{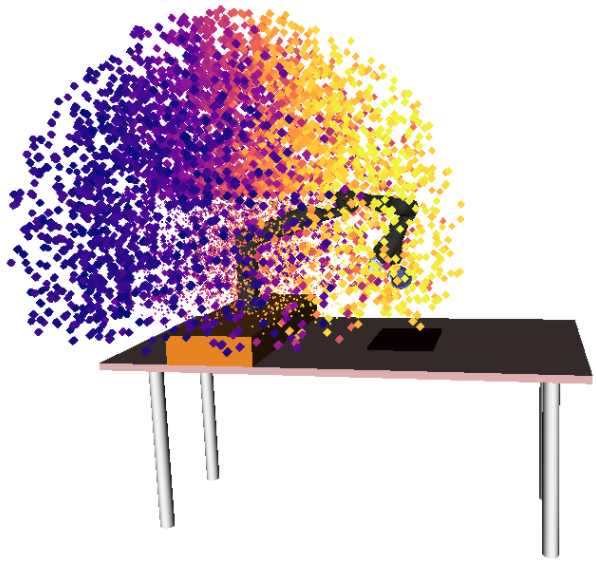}
\end{subfigure}
\begin{subfigure}{.26\textwidth}
  \centering
  \includegraphics[width=\textwidth,left]{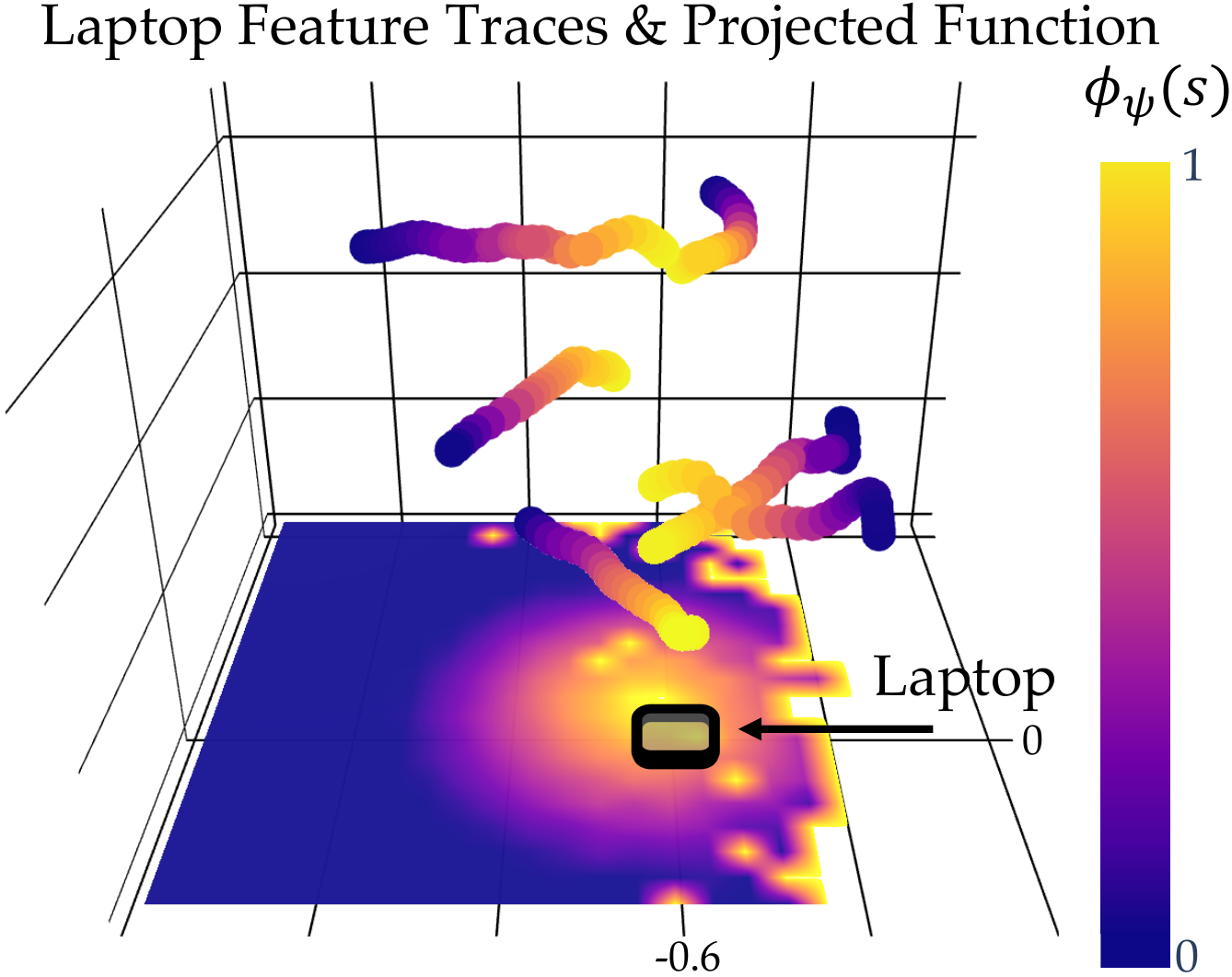}
\end{subfigure}
\caption{Visualization of the experimental setup, learned feature values $\phi_{\psi}(s)$, and training feature traces $\trace$ for \textit{table} (up) and \textit{laptop} (down). We display the feature values $\phi_{\psi}(\sysstate)$ for states $\sysstate$ sampled from the reachable set of the 7DoF arm, as well as their projections onto the $yz$ and $xy$ planes.} 
\label{fig:Laptop_Feature_Exp}
\vspace{-3mm}
\end{figure}

\subsubsection{Experimental Design.}
We conduct our experiments on a 7-DoF JACO robotic arm. We investigate six features in the context of personal robotics:
a) \textit{table}: distance of the \ac{EE} to the table;
b) \textit{coffee}: keeping the coffee cup upright;
c) \textit{laptop}: 0.3 meter $xy$-plane distance of the \ac{EE} to a laptop position, to avoid passing over the laptop at any height;
d) \textit{test laptop location}: same as \textit{laptop} but the test position differs from the training ones;
e) \textit{proxemics}: keeping the \ac{EE} away from the human, more so when moving in front of them, and less so when moving on their side;
f) \textit{between objects}: 0.2 meter $xy$-plane distance of the \ac{EE} to two objects -- the feature penalizes collision with objects, and, to a lesser extent, passing in between the objects. This feature requires some traces with explicit labels $l_0, l_n$. We approximate all features $\phi_{\psi}$ by neural networks (2 layers, 64 units each), and train them on a set of traces $\Xi$ using stochastic gradient descent (see App. \ref{app:FERL_implementation} for details). 

For each feature, we collected a set $\mathcal{F}$ of 20 feature traces (40 for the complex \textit{test laptop location} and \textit{between objects}) from which we sample subsets $\Xi \in \mathcal{F}$ for training. 
We determine for each feature what an informative and intuitive set of traces would be, i.e. how to choose the starting states to cover enough of the space (details in App. \ref{app:FERL_traces}). 
As described in Sec. \ref{sec:humaninput}, the human teaching the feature starts at a state where the feature is highly expressed, e.g. for \textit{laptop} that is the \ac{EE} positioned above the laptop. They then move the \ac{EE} away until the distance is equal to the desired radius. They do this for a few different directions and heights to give a diverse dataset.

Our raw state space consists of the 27D $xyz$ positions of all robot joints and objects in the scene, as well as the rotation matrix of the \ac{EE}. 
It was surprisingly difficult to train on both positions and orientations due to spurious correlations in the raw state space, hence we show results for training only on positions or only on orientations. This speaks to the need for methods that can handle correlated input spaces, which we expand on in App. \ref{app:rawstate}.

\paragraph{Manipulated Variables}

We manipulate the number of traces $N$ the learner gets access to. We want to see trends in how the quality of the learned features changes with more or less data available.

\paragraph{Dependent Measures}

After training a feature $\phi_{\psi}$, we measure error compared to the ground truth feature $\phi_{\text{True}}$ that the expert tries to teach, on a test set of states $\mathcal{S}_{\text{Test}}$.
To form $\mathcal{S}_{\text{Test}}$, we uniformly sample 10,000 states from the robot's reachable set. Importantly, many of these test points are far from the training traces, probing the generalization of the learned features $\phi_{\psi}$. We measure error via the \ac{MSE}, $\text{MSE} = \frac{1}{|\mathcal{S}_{\text{Test}}|} \sum_{\mathcal{S}_{\text{Test}}}||\phi_{\psi}(\sysstate) - \phi_{\text{True}}(\sysstate)||^2$. To ground the \ac{MSE} values, we normalize them with the mean \ac{MSE} of a randomly initialized untrained feature function, $\text{MSE}_{\text{norm}} = \frac{\text{MSE}}{\text{MSE}_{\text{random}}}$, hence a value of 1.0 equals random performance. For each number of feature traces $N$, we run 10 experiments sampling different traces from $\mathcal{F}$, and calculate $\text{MSE}_{\text{norm}}$.

\paragraph{Hypotheses} \hfill

\noindent\textbf{H1:} With enough data, FERL learns good features.

\noindent\textbf{H2:} FERL learns increasingly better features with more data.

\noindent\textbf{H3:} FERL becomes less input-sensitive with more data.

\begin{figure}
\includegraphics[width=0.47\textwidth]{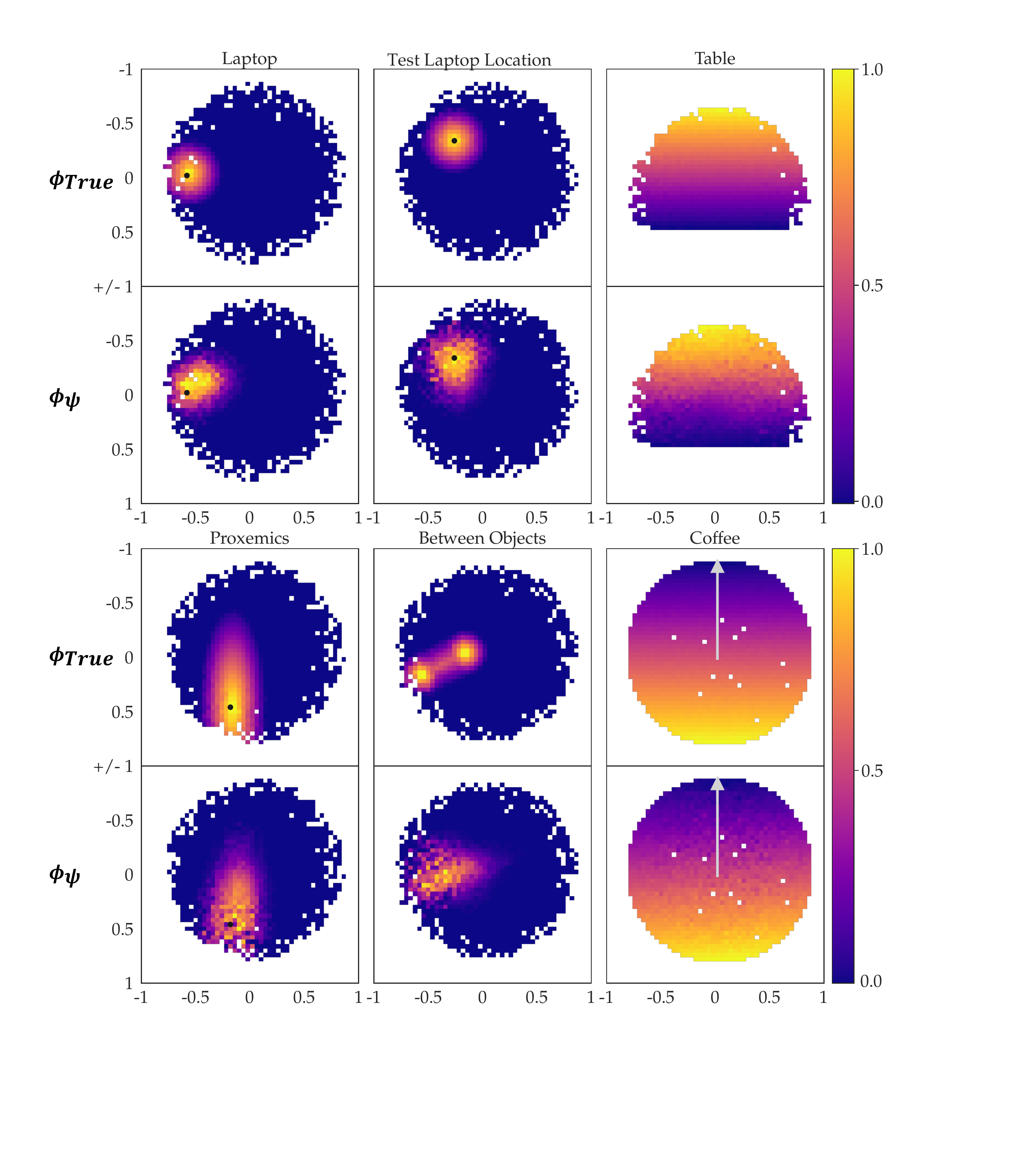}
\centering
\caption{The plots display the ground truth $\phi_{\text{True}}$ (top rows) and learned feature values $\phi_{\psi}$ (bottom rows) over $\mathcal{S}_{\text{Test}}$, averaged and projected onto a representative 2D subspace: the $xy$-plane, the $yz$-plane (table), and the $xz$ orientation plane for \textit{coffee} (the arrow represents the cup upright).}
\label{fig:FERL_Qual}
\vspace{-2mm}
\end{figure}

\subsubsection{Qualitative results.}
We first inspect the results qualitatively, for $N\!=\!10$. In Fig. \ref{fig:Laptop_Feature_Exp} we show the learned \textit{table} and \textit{laptop} features $\phi_{\psi}$ by visualizing the position of the \ac{EE} for all 10,000 points in our test set. The color of the points encodes the learned feature values $\phi_{\psi}(\sysstate)$ from low (blue) to high (yellow): \textit{table} is highest when the \ac{EE} is farthest, while \textit{laptop} peaks when the \ac{EE} is above the laptop.
In Fig. \ref{fig:FERL_Qual}, we illustrate the \ac{GT} feature values $\phi_{\text{True}}$ and the trained features $\phi_{\psi}$ by projecting the test points on 2D sub-spaces and plotting the average feature value per 2D grid point.
For Euclidean features we used the \ac{EE}'s \textit{xy}-plane or  \textit{yz}-plane (\textit{table}), and for \textit{coffee} we project the $x$-axis basis vector of the \ac{EE} after forward kinematic rotations onto the \textit{xz}-plane (arrow up represents the cup upright). White pixels are an artifact of sampling.

We observe that $\phi_{\psi}$ resembles $\phi_{\text{True}}$ very well for most features. Our most complex feature, \textit{between objects}, does not recreate the \ac{GT} as well, although it does learn the general shape. 
However, we note in App. \ref{app:betweenobjects} that in smaller raw input space it is able to learn the fine-grained GT structure.
This implies that spurious correlation in input space is a problem, hence for complex features more data or active learning methods to collect informative traces are required.

\subsubsection{Quantitative analysis.} 
\begin{figure}
\includegraphics[width=0.47\textwidth]{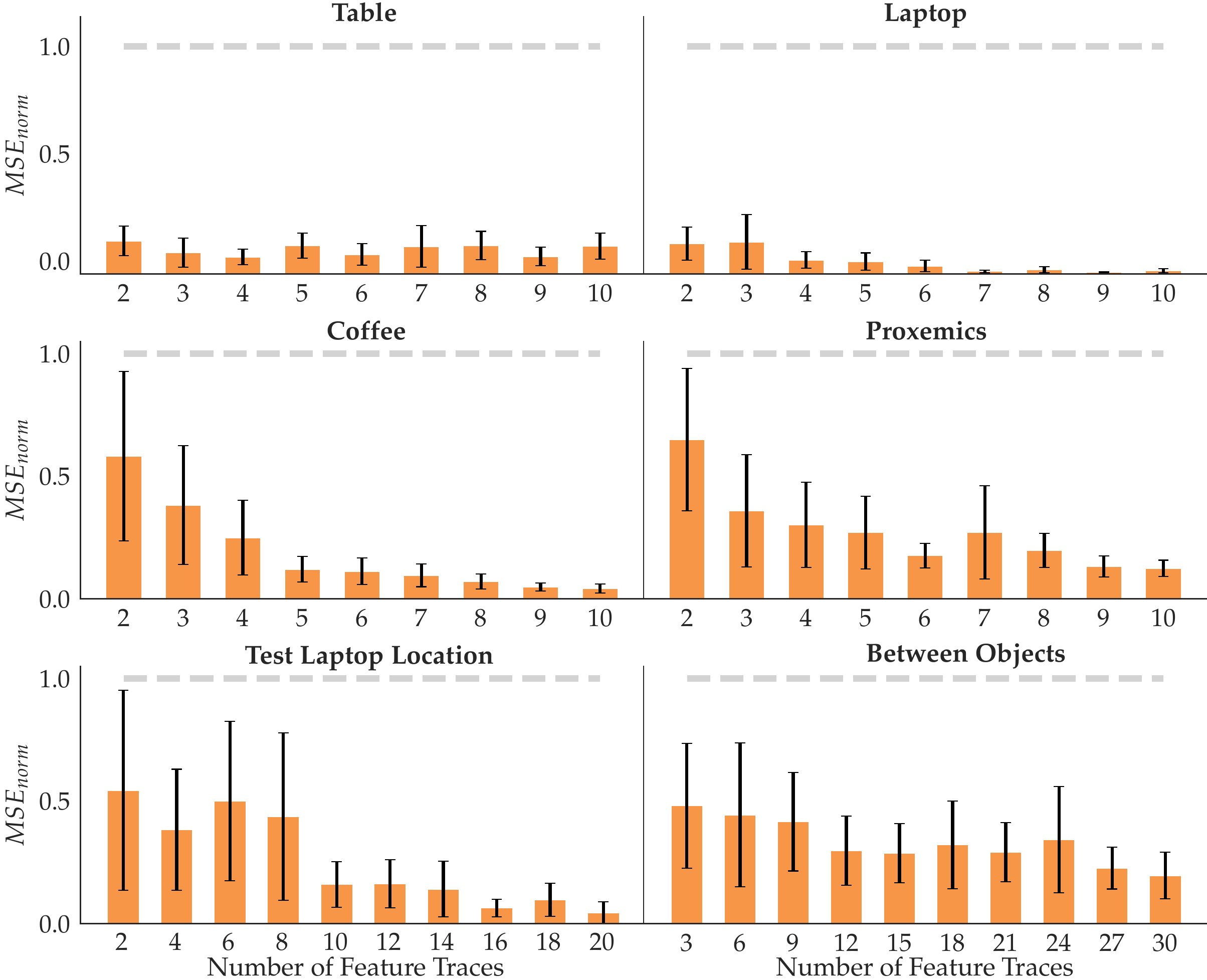}
\centering
\caption{For each feature, we show the $\text{MSE}_{\text{norm}}$ mean and standard error across 10 random seeds with an increasing number of traces (orange) compared to random (gray).}
\label{fig:FERL_MSE}
\vspace{-5mm}
\end{figure}

Fig. \ref{fig:FERL_MSE} displays the means and standard errors across 10 seeds for each feature with data increase. To test H1, we look at the errors with the maximum amount of data. Indeed, FERL achieves small errors, put in context by the comparison with the error a random feature incurs (gray line). This is confirmed by an ANOVA with random vs. FERL as a factor and the feature ID as a covariate, finding a significant main effect ($F(1,113)=372.0123,p<.0001$). In line with H2, most features have decreasing error with increasing data. Indeed, an ANOVA with amount of data as a factor and feature ID as a covariate found a significant main effect ($F(8,526)=21.1407,p<.0001$). Lastly, in line with H3, we see that the standard error on the mean decreases when FERL gets more data. To test this, we ran an ANOVA with the standard error as the dependent measure and the amount of data as a factor, finding a significant main effect ($F(8,45)=3.098,p=.0072$).
In summary, the qualitative and quantitative results support our hypotheses and suggest that our method requires few traces to reliably learn feature functions $\phi_{\psi}$ that generalize well to states not seen during training.

\subsection{Reward Learning}
\label{sec:reward_experts}

We compare \ac{FERL} for reward learning to a deep \ac{IRL} baseline.

\subsubsection{Experimental Setup.} We run experiments on the robot arm in three settings in which two features are known ($\phi_{\text{coffee}}, \phi_{\text{known}}$) and one feature is unknown. In all tasks, the true reward is $r_{\text{true}}\!=\!(0, 10, 10)(\phi_{\text{coffee}}, \phi_{\text{known}}, \phi_{\text{unknown}})^T$.
We include $\phi_{\text{coffee}}$ with zero weight to evaluate if the methods can learn to ignore irrelevant features.
In task 1, $\phi_{\text{laptop}}$ is unknown and the known feature is $\phi_{\text{table}}$; in task 2, $\phi_{\text{table}}$ is unknown and $\phi_{\text{laptop}}$ is known; and in task 3, $\phi_{\text{proxemics}}$ is unknown and $\phi_{\text{table}}$ is known.

\paragraph{Manipulated Variables} We manipulated the \textit{learning method} with 2 levels: \ac{FERL} and an adapted \ac{ME-IRL} baseline \cite{maxent,finn2016gcl, wulfmeier2016maxentirl} learning a deep reward function from demonstrations. We model the \ac{ME-IRL} reward function $r_{\omega}$ as a neural network with 2 layers, 128 units each. For a fair comparison, we gave $r_{\omega}$ access to the known features: once the 27D Euclidean input is mapped to a final neuron, a last layer combines it with the known feature vector.

Also for a fair comparison, we took great care to collect a set of demonstrations for \ac{ME-IRL} designed to be as informative as possible: we chose diverse start and goal configurations for the demonstrations, and focused some of them on the unknown feature and some on learning a combination between features (see App. \ref{app:MEIRL_demos}). 
Moreover, \ac{FERL} and \ac{ME-IRL} rely on different input types: \ac{FERL} on feature traces $\trace$ and pushes $a_H$ and \ac{ME-IRL} on a set of near-optimal demonstrations $\mathcal{D}^*$. To level the amount of data each method has access to, we collected the traces $\Xi$ and demonstrations $\mathcal{D}^*$ such that \ac{ME-IRL} has more data points: the average number of states per demonstration/trace were 61 and 39, respectively.


The gradient of the \ac{ME-IRL} objective with respect to the reward parameters $\omega$ can be estimated by: $\nabla_{\omega}\mathcal{L} \!\approx\! \frac{1}{|\mathcal{D}^*|}\sum_{\tau \in \mathcal{D}^*} \!\!\nabla_{\omega}R_{\omega}(\tau) \!-\! \frac{1}{|\mathcal{D}^{\omega}|}\sum_{\tau \in \mathcal{D}^{\omega}}\!\! \nabla_{\omega}R_{\omega}(\tau)$ \cite{wulfmeier2016maxentirl,finn2016gcl}. 
Here, $R_{\omega}(\tau)\!=\!\sum_{s \in \tau}\!r_{\omega}(\sysstate)$ is the parametrized reward, $\mathcal{D}^*$ the expert demonstrations, and $\mathcal{D}^{\omega}$ are trajectory samples from the $r_{\omega}$ induced near optimal policy.
We use TrajOpt \cite{schulman2013trajopt} to obtain the samples $\mathcal{D}^{\omega}$ (see App. \ref{app:MEIRL_implementation} for details).

\paragraph{Dependent Measures}

We compare the two reward learning methods across two metrics commonly used in the \ac{IRL} literature~\cite{choi2011inverse}: 1) \textit{Reward Accuracy}: how close to \ac{GT} the learned reward is, and 2) \textit{Behavior Accuracy}: how well do the behaviors induced by the learned rewards compare to the \ac{GT} optimal behavior, measured by evaluating the induced trajectories on \ac{GT} reward. 
For \textit{Reward Accuracy}, we vary the number of traces / demonstrations each learner gets access to, and measure the \ac{MSE} compared to the \ac{GT} reward on $\mathcal{S}_{\text{Test}}$, similar to Sec. \ref{sec:feature_expert}. For \textit{Behavior Accuracy}, we train \ac{FERL} and \ac{ME-IRL} with a set of 10 traces / demonstrations. 
We then use TrajOpt~\cite{trajopt} to produce optimal trajectories for 100 randomly selected start-goal pairs under the learned rewards. We evaluate the trajectories with the \ac{GT} reward $r_{\text{true}}$ and divide by the reward of the \ac{GT} induced trajectory for easy relative comparison.

\paragraph{Hypotheses} \hfill

\noindent\textbf{H4:} \ac{FERL} learns rewards that better generalize to the state space than \ac{ME-IRL}. 

\noindent\textbf{H5:} \ac{FERL} performance is less input-sensitive than \ac{ME-IRL}.

\begin{figure}
\includegraphics[width=.47\textwidth]{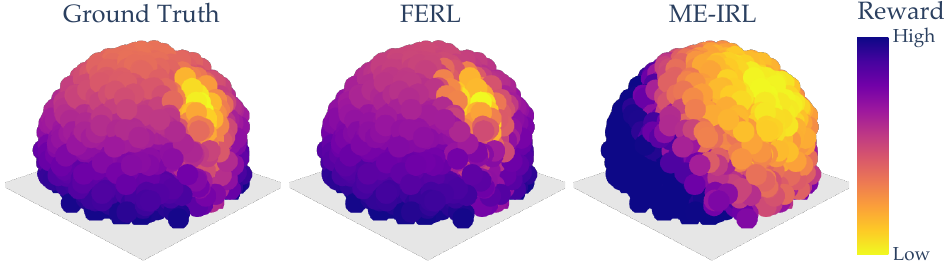}
\centering
\caption{Visual comparison of the ground truth, \ac{FERL}, and \ac{ME-IRL} rewards for task 1.} 
\label{fig:VisualComparison}
\vspace{-5mm}
\end{figure}

\subsubsection{Qualitative Comparison.} In Fig. \ref{fig:VisualComparison}, we show the learned \ac{FERL} and \ac{ME-IRL} rewards as well as the \ac{GT} for task 1 evaluated at the test points (see App. \ref{app:cases23} for tasks 2 and 3). 
As we can see, by first learning the \textit{laptop} feature and then the reward on the extended feature vector, \ac{FERL} is able to learn a fine-grained reward structure closely resembling GT. Meanwhile, \ac{ME-IRL} learns some structure capturing where the laptop is, but not enough to result in a good trade-off between the active features. 

\subsubsection{Quantitative Analysis.} 
\begin{figure}
\begin{minipage}[t]{.47\textwidth}
  \includegraphics[width=\textwidth,left]{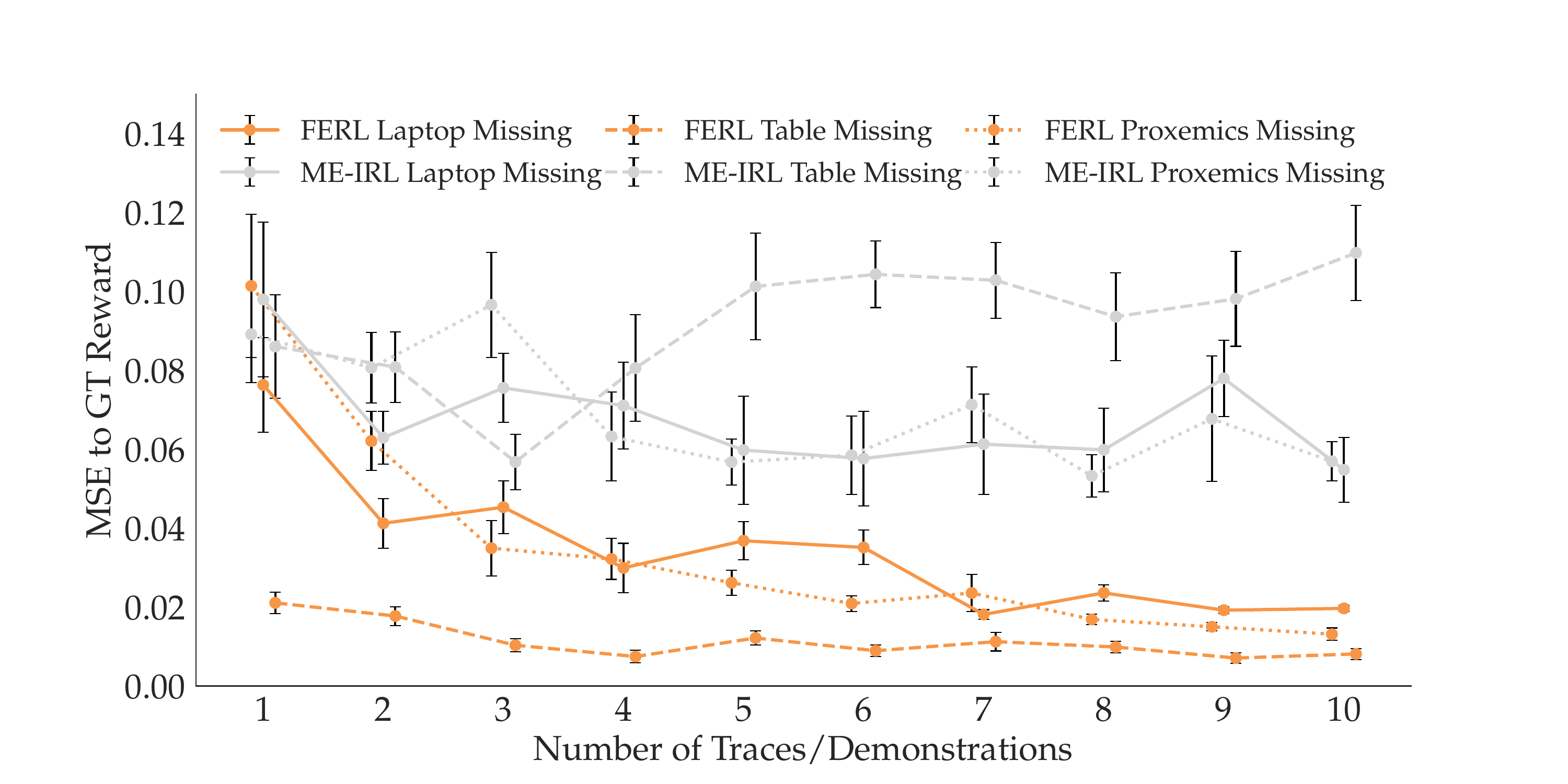}
   \caption{MSE of FERL and ME-IRL to \ac{GT} reward.}
   \label{fig:MSEComparison}
\end{minipage}%
\hfill
\begin{minipage}[t]{.37\textwidth}
  \includegraphics[width=\textwidth,right]{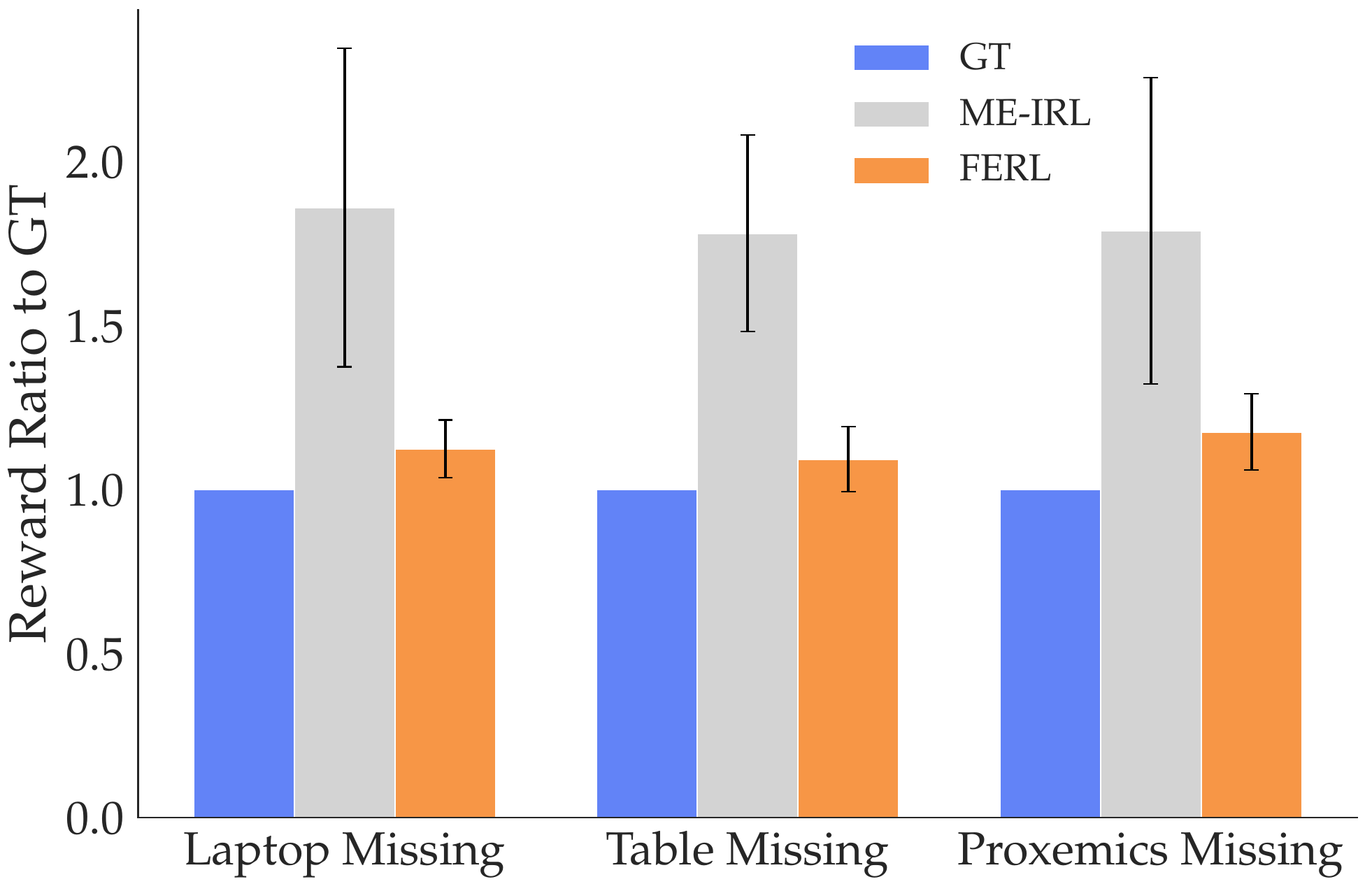}
   \caption{Induced trajectories' reward ratio.}
   \label{fig:InducedTrajectories}
\end{minipage}
\vspace{-3mm}
\end{figure}

To compare \textit{Reward Accuracy}, we show in Fig. \ref{fig:MSEComparison} the \ac{MSE} mean and standard error across 10 seeds, with increasing training data. We visualize results from all 3 tasks, with \ac{FERL} in orange and \ac{ME-IRL} in gray. \ac{FERL} is closer to GT than \ac{ME-IRL} no matter the amount of data, supporting H4. To test this, we ran an ANOVA with learning method as the factor, and with the task and data amount as covariates, and found a significant main effect (F(1, 595) = 335.5253, p < .0001).

Additionally, the consistently decreasing MSE in Fig. \ref{fig:MSEComparison} for \ac{FERL} suggests that our method gets better with more data; in contrast, the same trend is inexistent with \ac{ME-IRL}. Supporting H5, the high standard error that \ac{ME-IRL} displays implies that it is highly sensitive to the demonstrations provided and the learned reward likely overfits to the expert demonstrations. We ran an ANOVA with  standard error as the dependent measure, focusing on the $N=10$ trials which provide the maximum data to each method, with the learning method as the factor and the task as a covariate. We found that the learning method has a significant effect on the standard error (F(1, 4) = 12.1027, p = .0254). With even more data, this shortcoming of IRL might disappear; however, this would pose an additional burden on the human, which our method successfully alleviates. 

Lastly, we looked at \textit{Behavior Accuracy} for the two methods. Fig. \ref{fig:InducedTrajectories} illustrates the reward ratios to \ac{GT} for all three tasks. The \ac{GT} ratio is 1 by default, and the closer to 1 the ratios are, the better the performance because all rewards are negative. The figure further supports H4, showing that \ac{FERL} rewards produce trajectories that are preferred under the \ac{GT} reward over \ac{ME-IRL} reward trajectories. An ANOVA using the task as a coviarate reveals a significant main effect for the learning method (F(1, 596) = 14.9816, p = .0001).


%% file: 4_FERL_users.tex
\section{FERL User Study}
\label{sec:FERL_users}

\begin{figure*}
\centering
\begin{subfigure}{.33\textwidth}
  \centering
  \includegraphics[width=\textwidth,left]{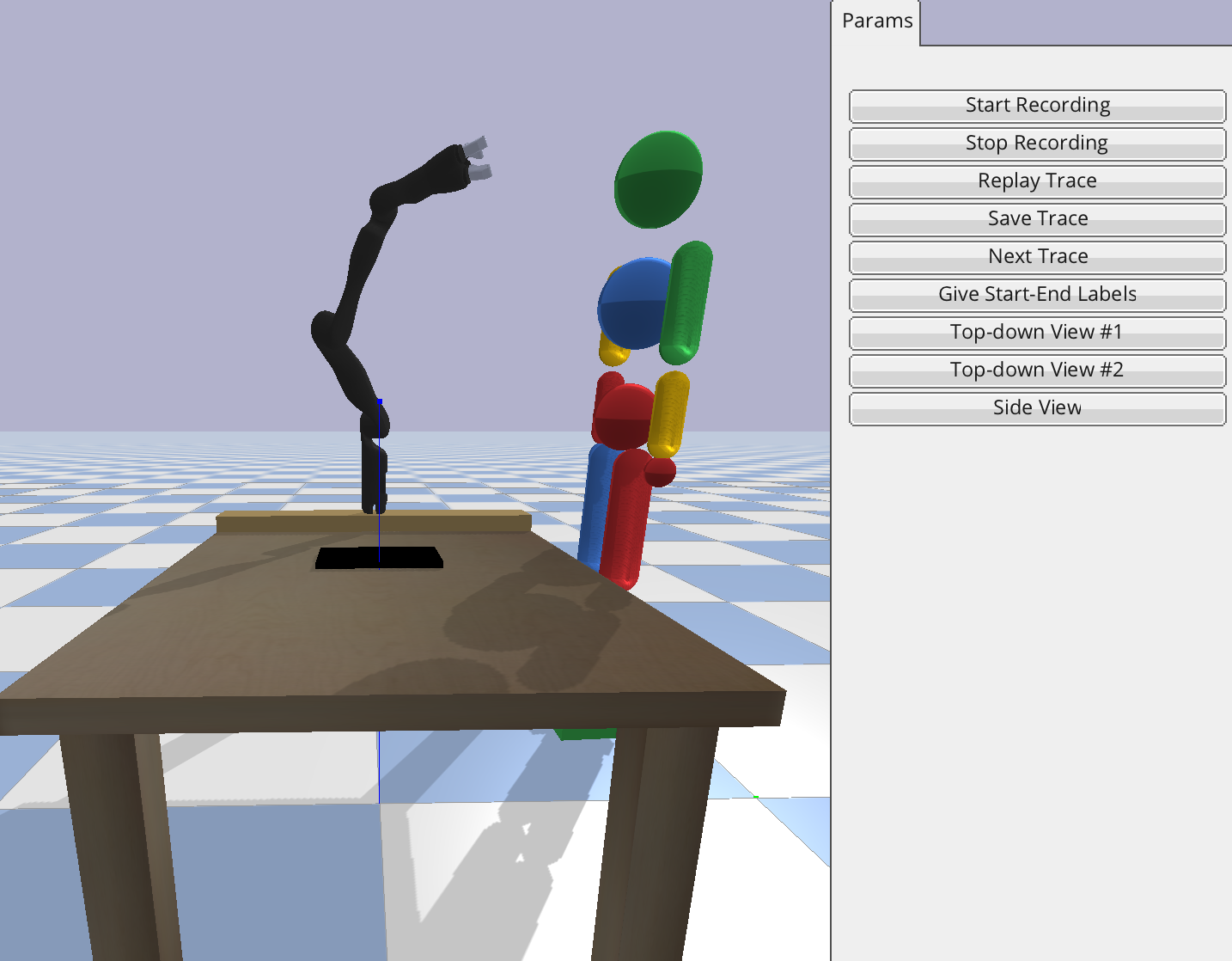}
\end{subfigure}%
\begin{subfigure}{.3\textwidth}
  \centering
  \includegraphics[width=\textwidth,left]{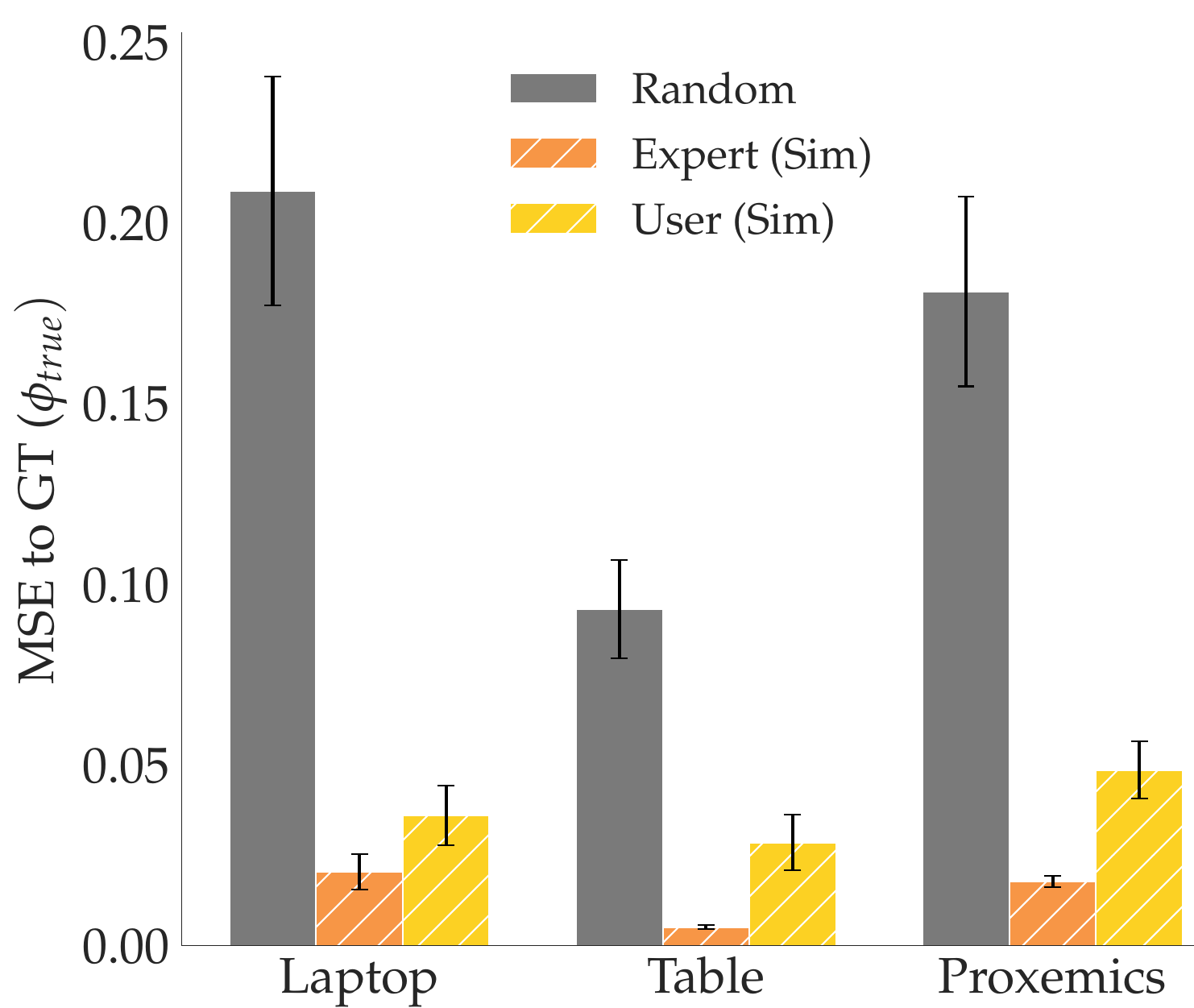}
\end{subfigure}
\begin{subfigure}{.3\textwidth}
  \centering
  \includegraphics[width=\textwidth,left]{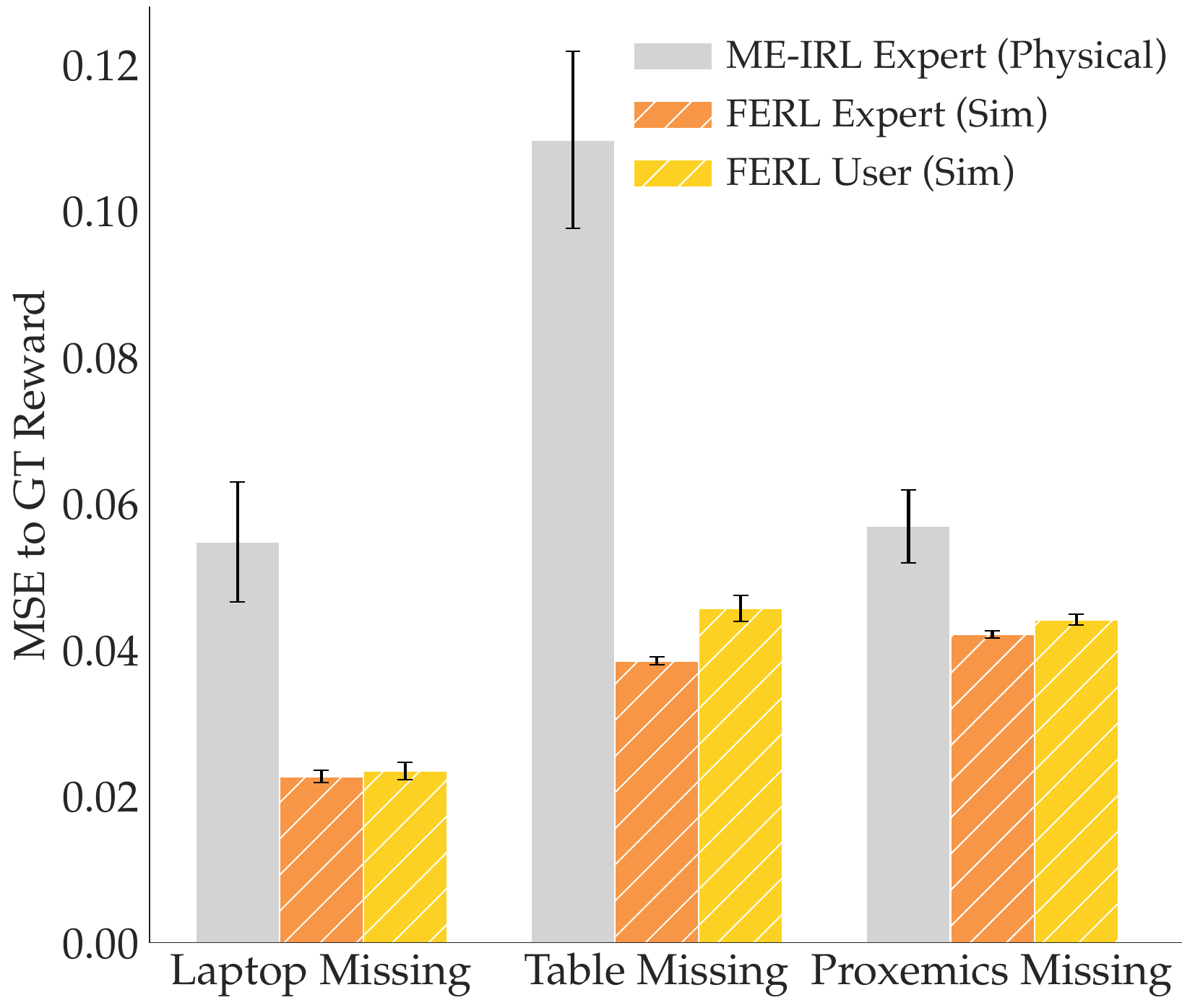}
\end{subfigure}
\caption{The user study simulator interface (left). MSE to GT feature comparing expert and non-expert users (middle). MSE to GT reward for ME-IRL with physical demonstrations and FERL with user and expert features learned in simulation (right).}
\label{fig:user_study}
\end{figure*}

In Sec. \ref{sec:FERL_experts}, we analyzed \ac{FERL}'s efficacy with expert data. We now design a user study to test how well non-expert users can teach features with \ac{FERL} and how easily they can use the \ac{FERL} protocol. 

\subsection{Feature Learning}
\label{sec:feature_users}

We first collect user feature traces to learn \ac{FERL} features, and we later test their performance in reward learning in Sec. \ref{sec:reward_users}.

\subsubsection{Experimental Design}
Due to COVID, we replicated our set-up from Fig. \ref{fig:front_fig} in a pybullet simulator~\cite{coumans2019} in which users can move a 7 DoF JACO robotic arm using their cursor. Through the interface in Fig. \ref{fig:user_study} (left), the users can drag the robot to provide feature traces, and use the buttons for recording, saving, and discarding them.

The user study is split into two phases: familiarization and teaching. In the first phase, we introduce the user to the task context, the simulation interface, and how to provide feature traces through an instruction video and a manual. Next, we describe and 3D visualize the familiarization task feature \textit{human} (0.3 meter $xy$-plane distance of the \ac{EE} to the human position), after which we ask them to provide 10 feature traces to teach it. Lastly, we give the users a chance to see what they did well and learn from their mistakes by showing them a 3D visualization of their traces and the learned feature.

In the second phase, we ask users to teach the robot the three features from Sec. \ref{sec:reward_experts}: \textit{table}, \textit{laptop}, and \textit{proxemics}. This time, we don't show the learned features until after all three tasks are complete.

\paragraph{Manipulated Variables}
We manipulate the \textit{input type} with three levels: random, expert, and user. For random, we randomly initialize 12 feature functions per task; for expert, the authors collected 20 traces per task in the simulator, then subsampled 12 sets of 10 that lead to features of similar MSEs to the ones in the physical setup before;
for user, each person provided 10 traces per task.

\paragraph{Dependent Measures}
Our objective metric is the learned feature's \ac{MSE} compared to the \ac{GT} feature on $\mathcal{S}_{\text{Test}}$, similar to Sec. \ref{sec:FERL_experts}. Additionally, to assess the users’ interaction experience we administered the subjective 7-point Likert scale survey from Fig. \ref{fig:study_qual}, with some items inspired by NASA-TLX~\cite{HART1988NASATLX}. After they provide the feature traces for all 3 tasks, we ask the top eight questions in Fig. \ref{fig:study_qual}. The participants then see the 3D visualizations of their feature traces and learned features, and we survey all 11 questions as in Fig. \ref{fig:study_qual} to see if their assessment changed.




\paragraph{Participants}
We recruited 12 users (11 male, aged 18-30) from the campus community to interact with our simulated JACO robot and provide feature traces for the three tasks. All users had technical background, so we caution that our results will speak to FERL's usability with this population rather than the general population.

\paragraph{Hypotheses} Using our objective and subjective metrics, we test:  

\noindent\textbf{H6:} FERL learns good features even with non-expert user data. 

\noindent\textbf{H7:} Users find it easy to think of traces to give the robot, believe they understand how these traces influence the learned feature, believe they were successful teachers, and find our teaching protocol intuitive (little mental/physical effort, time, or stress).

\subsubsection{Analysis}

\paragraph{Objective} Fig. \ref{fig:user_study} (middle) summarizes the results by showing how the \ac{MSE} varies with each of our input types, for each task feature. Right off the bat, we notice that in line with H6, the MSEs for the user features are much closer to the expert level than to random. We ran an ANOVA with input type as a factor and task as a covariate, finding a significant main effect (F(2, 103) = 132.7505, p < .0001). We then ran a Tukey HSD post-hoc, which showed that the MSE for random input was significantly higher than both expert (p < .0001) and user (p < .0001), and found no significant difference between expert and user (p = .0964).
While this does not mean that user features are as good as expert features (we expect some degradation in performance when going to non-experts), it shows that they are substantially closer to them than to random, i.e. the user features maintain a lot of signal despite this degradation.

\begin{figure}
\includegraphics[width=.47\textwidth]{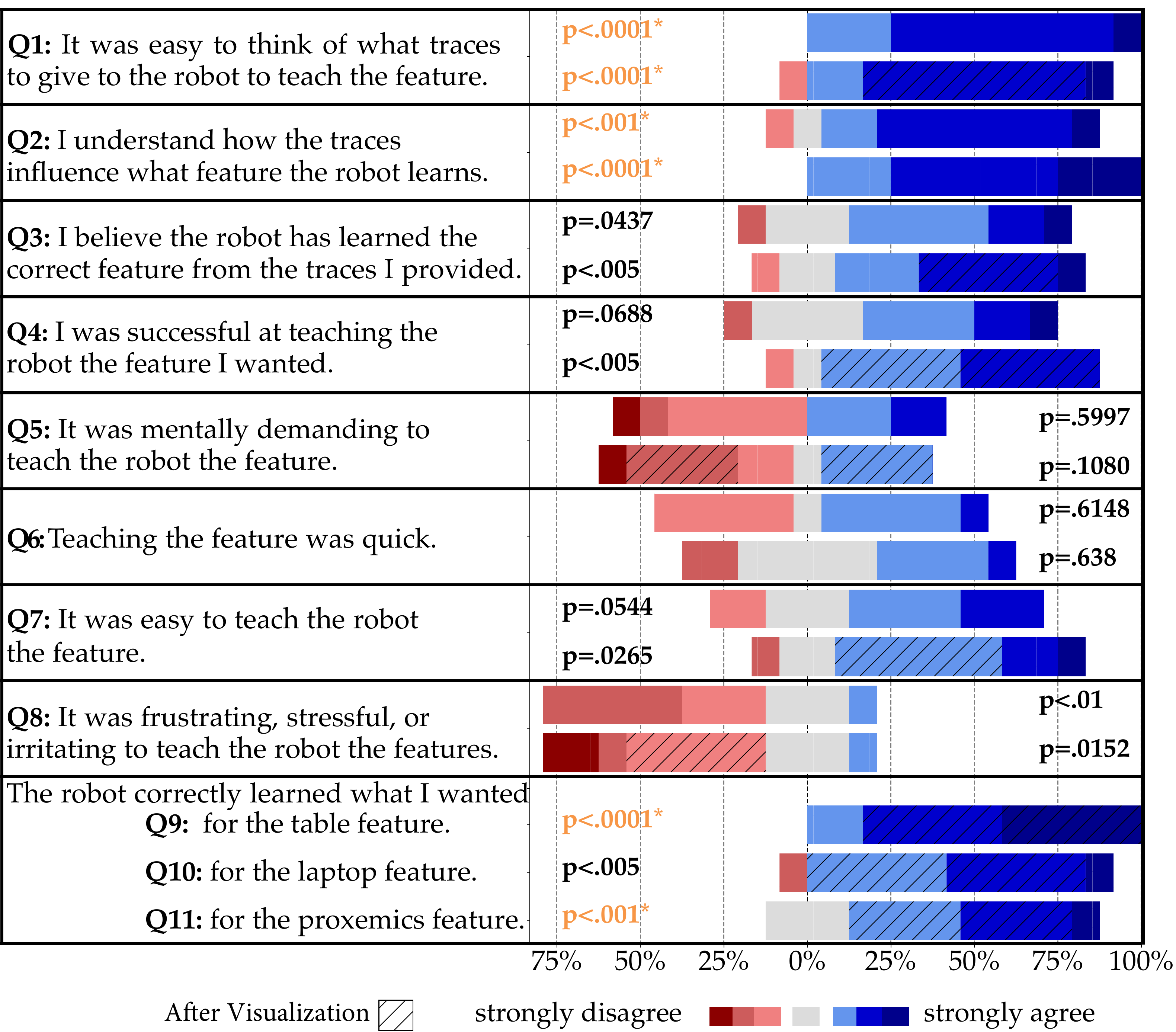}
\centering
\caption{Questions, answer distributions, and p-values 
(2-sided t-test against the middle score 4) from the user study. The p-values in orange are significant after adjusted for multiple comparisons using the Bonferroni correction.}
\label{fig:study_qual}
\end{figure}

\paragraph{Subjective} In Fig. \ref{fig:study_qual}, we see the Likert survey scores before and after the users saw the teaching results. For every question, we report 2-sided t-tests against the neutral score 4. These results are in line with H7, although the evidence for finding the teaching protocol intuitive is weaker, and participants might have a bias to be positive given they are in a study.
In fact, several participants mentioned in their additional remarks that they had a good idea of what traces to give, and the only frustrating part was the GUI interface, which was necessary because in-person studies are not possible during the COVID pandemic
("I know what it wants, but the interface makes it difficult to give those exact traces"); performing the experiment as it was originally intended with the real robot arm would have potentially alleviated this issue ("With manual control of the arm it would have been a lot easier.").

Looking before and after the visualization, we notice a trend: seeing the result seems to reinforce people’s belief that they were effective teachers (Q3, Q4), also noticed in their comments 
("Surprised that with limited coverage, it generalized pretty well."). Additionally, in support of H6, we see significant evidence that users thought the robot learned the correct feature (Q9-Q11).

Lastly, we wanted to know if there was a correlation between subjective scores and objective performance. We isolated the ``good teachers'' -- the participants who scored better than average on all 3 feature tasks in the objective metric, and compared their subjective scores to the rest of the teachers. By running a factorial likelihood-ratio test for each question, we found a significant main effect for good teachers: they are more certain that the robot has learned a correct feature even before seeing the results (Q3, p = .001), are more inclined to think they were successful (Q4, p = .0203), and find it significantly easier to teach features (Q7, p = .0202).

\subsection{Reward Learning}
\label{sec:reward_users}

The objective results in the previous section show that while users' performance degrades from expert performance, they are still able to teach features with a lot of signal. 
We now want to test how important the user-expert feature quality gap is when it comes to using these features for reward learning.

\subsubsection{Experimental Design}
We take the features learned from the user study, and perform reward learning with them.

\paragraph{Manipulated Variables}
We manipulate the \textit{learning method}, \ac{FERL} or \ac{ME-IRL}, just like in Sec. \ref{sec:reward_experts}. Because corrections and demonstrations would be very difficult in simulation, we use for ME-IRL the expert data from the physical robot. For \ac{FERL}, we use the user data from the simulation, and the expert corrections that teach the robot how to combine the learned feature with the known ones. Note that this gives \ac{ME-IRL} an advantage, since its data is both generated by an expert, and on the physical robot. Nonetheless, we hypothesize that the advantage of the divide-and-conquer approach is stronger.


\paragraph{Dependent Measures} We use the same objective metric as \textit{Reward Accuracy} in the expert comparison in Sec. \ref{sec:reward_experts}: the learned reward \ac{MSE} to the \ac{GT} reward on $\mathcal{S}_{\text{Test}}$.

\paragraph{Hypothesis}
\textbf{H8:} FERL learns more generalizable rewards than ME-IRL even when using features learned from data provided by non-experts in simulation.

\subsubsection{Analysis}

Fig. \ref{fig:user_study} (right) illustrates our findings for the reward comparison. In the figure, we added \ac{FERL} with expert-taught simulation features for reference: we subsampled sets of 10 from 20 expert traces and trained 12 expert features for each of our 3 task features. We see that, \textit{even though \ac{ME-IRL} was given the advantage of using physical expert demonstrations, it still severely underperforms when compared to FERL with both expert and user features learned in simulation.} This finding is crucial because it underlines the power of our divide-and-conquer approach in reward learning: even when given imperfect features, the learned reward is superior to trying to learn everything implicitly from demonstrations.

We verified the significance of this result with an ANOVA with the learning method as a factor and the task as a covariate. We found a significant main effect for the learning method (F(1, 62) = 41.2477, p < .0001), supporting our H8.

%% file: 5_discussion.tex
\section{Discussion}
\label{sec:discussion}

\change{In this work, we proposed \ac{FERL}, a framework for learning rewards when the initial feature set cannot capture human preferences. We introduced feature traces as human input for learning features from raw state. In experiments and a user study, we showed that \ac{FERL} outperforms deep reward learning from demonstrations (\ac{ME-IRL}) in data-efficiency, generalization, and sensitivity to input data.}

\paragraph{Potential Implications for Learning Complex Rewards from Demonstrations.}
Reward learning from raw state space with expressive function approximators is considered difficult because there exists a large set of functions $r_\theta(\sysstate)$ that could explain the human input, e.g. for demonstrations many functions $r_\theta(\sysstate)$ induce policies that match the demonstrations' state expectation. The higher dimensional the state $\sysstate \in \mathbb{R}^d$, the more human input is needed to disambiguate between those functions sufficiently to find a reward $r_\theta$ which accurately captures human preferences and thereby generalizes to states not seen during training and not just replicates the demonstrations' state expectations as in \ac{IRL}.
We are hopeful that our method of collecting feature traces rather than just demonstrations has potential implications broadly for non-linear (deep) reward learning, as a way to better disambiguate the reward and improve generalization.

While in this paper we focused on adapting a reward online, we also envision our method used as part of a "divide-and-conquer" alternative to \ac{IRL}: collect feature traces for the salient non-linear criteria of the reward, then use demonstrations to figure out how to combine them. The reason this might help relative to relying on demonstrations for everything is that demonstrations aggregate a lot of information. First, by learning features, we can isolate learning what matters from learning how to trade off what matters into a single value (the features vs. their combination) -- in contrast, demonstrations have to teach the robot about both at once. Second, feature traces give information about states that are not on optimal trajectories, be it states with high feature values that are undesirable, or states with low feature values where other, more important features have high values. 
Third, feature traces are also structured by the monotonicity assumption: they tell us relative feature values of the states along a trace, whereas demonstrations only tell us about the aggregate reward across a trajectory. These might be the reasons for the result in Fig. \ref{fig:MSEComparison}, \ref{fig:InducedTrajectories}, \ref{fig:user_study} (right), where the \ac{FERL} reward generalized better to new states than the demonstration-only IRL.

\paragraph{Limitations and Future Work.}
Due to the current pandemic, our user study is in a simulated environment instead of in person with the real robot, and most of our users had technical background. While our study still provides evidence that non-expert users can use \ac{FERL} to teach good features, it is unclear how people without technical background would perform. Further, we only tested whether users could teach features we tell them about, so we still need to test whether users can teach features they implicitly know about (as would happen when intervening to correct the robot).

\change{Even if people know the missing feature, it might be so abstract (e.g. comfort) that they would not know how to teach it.
Moreover, with the current feature learning protocol, they might find it cumbersome to teach discontinuous features like constraints. 
We could ease the human supervision burden by developing an active learning approach where the robot autonomously picks starting states most likely to result in informative feature traces. 
But for such complex features, it may be more effective to investigate combining feature traces with other types of structured human input.}

Lastly, while we show that \ac{FERL} works reliably in 27D, we want to extend it
to higher dimensional state spaces\change{, like images. 
Approaches that encode these spaces to lower dimensional representations or techniques from causal learning, such as Invariant Risk Minimization~\cite{Arjovsky2019InvariantRM}, could help tackle these challenges.}

%% file: 6_supplement.tex
\section{Experimental Details}

\subsection{Protocols for Feature Trace Collection}
\label{app:FERL_traces}

\begin{figure*}
\centering
\begin{subfigure}{.27\textwidth}
  \centering
  \includegraphics[width=\textwidth,left]{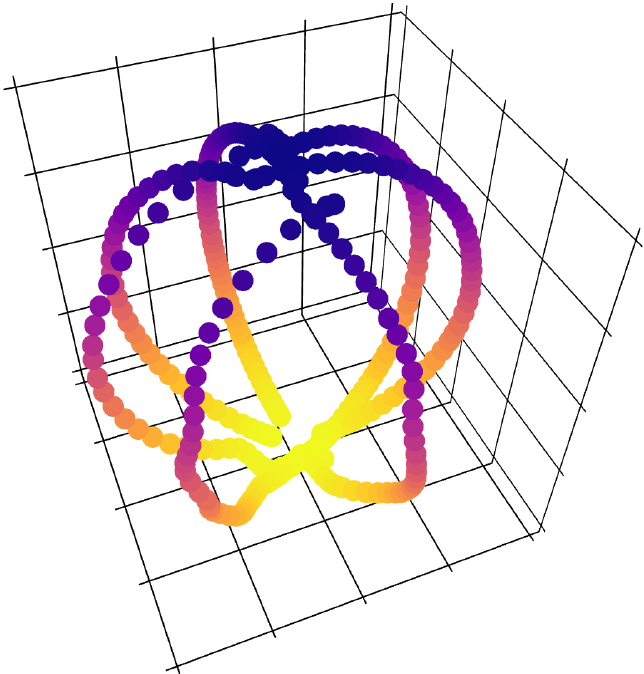}
\end{subfigure}%
\begin{subfigure}{.39\textwidth}
  \centering
  \includegraphics[width=\textwidth,left]{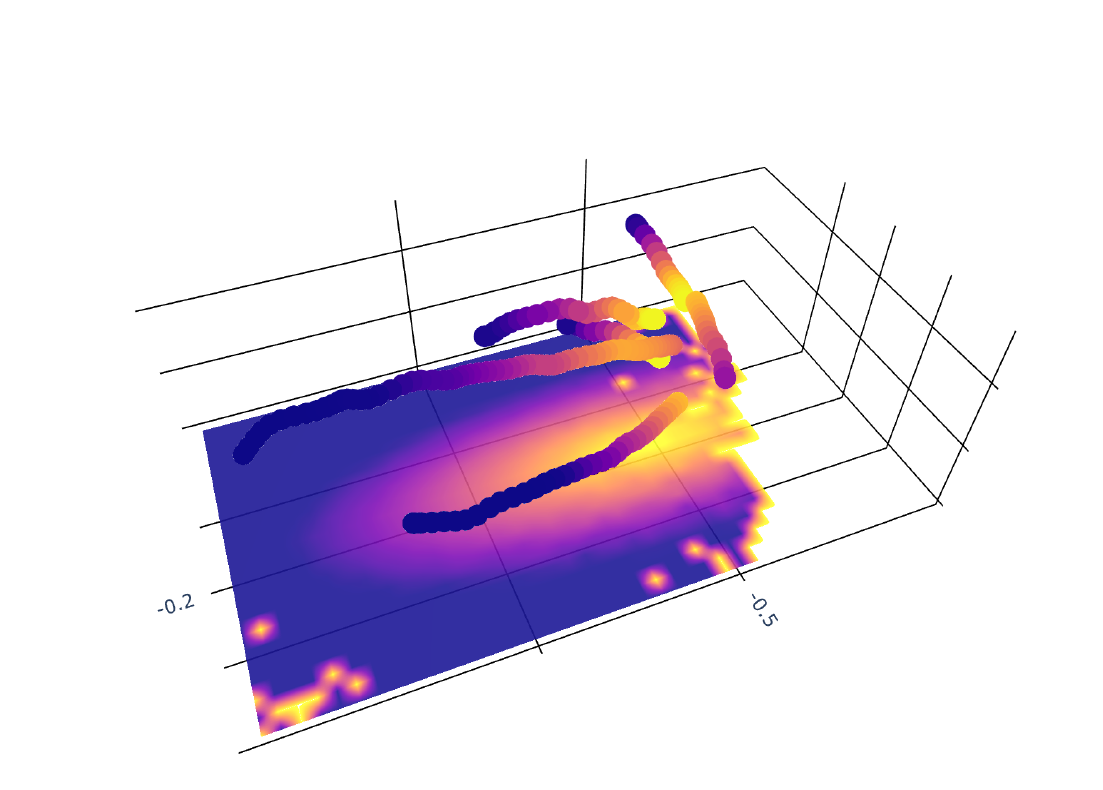}
\end{subfigure}
\begin{subfigure}{.33\textwidth}
  \centering
  \includegraphics[width=\textwidth,left]{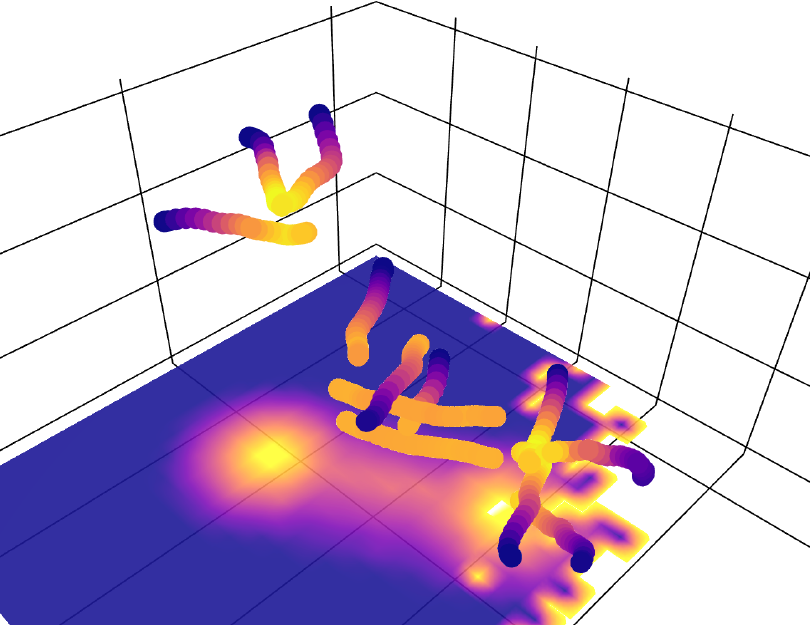}
\end{subfigure}
\caption{(Left) Feature traces for \textit{coffee}. We show the $xyz$ values of the $x$-axis base vector of the \acf{EE} orientation. The traces start with the \ac{EE} pointing downwards and move it upwards. (Middle) Feature traces for \textit{proxemics} with the human at $xy=[-0.2, -0.5]$, with \ac{GT} feature values projected on the $xy$-plane. (Right) Feature traces for \textit{between objects}.}
\label{fig:traces}
\end{figure*}

In this section, we present our protocol for collecting feature traces for the six features discussed in Sec. \ref{sec:feature_expert}. As we will see, the traces collected from the human only noisily satisfy the assumptions in Sec. \ref{sec:humaninput}. Nevertheless, as we showed in Sec. \ref{sec:feature_expert}, \ac{FERL} is able to learn high quality feature functions.

For \textit{table}, the person teaches that being close to the table, anywhere on the $xy$ plane, is desirable, whereas being far away in height is undesirable. As such, in Fig. \ref{fig:Laptop_Feature_Exp} on the left traces traverse the space from up at a height, until reaching the table. A few different starting configurations are helpful, not necessarily to cover the whole state space, but rather to have signal in the data: having the same trace 10 times would not be different from having it once.

For \textit{laptop}, as described in the text and shown in Fig. \ref{fig:Laptop_Feature_Exp} on the right, the person starts in the middle of the laptop, and moves away a distance equal to the bump radius desired. Having traces from a few different directions and heights helps learn a more distinct feature. For \textit{test laptop location}, the laptop's location at test time is not seen during training. Thus, the training traces should happen with various laptop positions, also starting in the middle and moving away as much distance as desired. 

When teaching the robot to keep the cup upright (\textit{coffee}), the person starts their traces by placing the robot in a position where the cup is upside-down, then moving the arm or rotating the \acf{EE} such that it points upright. Doing this for a few different start configurations helps.
Fig. \ref{fig:traces} (left) shows example traces colored with the true feature values.

When learning \textit{proxemics}, the goal is to keep the \ac{EE} away from the human, more so when moving in front of their face, and less so when moving on their side. As such, when teaching this feature, the person places the robot right in front of the human, then moves it away until hitting the contour of some desired imaginary ellipsis: moving further in front of the human, and not as far to the sides, in a few directions. Fig. \ref{fig:traces} (middle) shows example traces colored with the \acf{GT} feature values.

Lastly, for \textit{between objects} there are a few types of traces, all shown in Fig. \ref{fig:traces} (right). First, to teach a high feature value on top of the objects, some traces need to start on top of them and move away radially. Next, the person has a few options: 1) record a few traces spanning the line between the objects, at different heights, and labeling the start and the end the same; 2) starting anywhere on the imaginary line between the objects and moving perpendicularly away the desired distance, and labeling the start; 3) starting on top of one of the objects, moving towards the other then turning away in the direction orthogonal to the line between the objects.

\subsection{Protocols for Demonstration Collection}
\label{app:MEIRL_demos}

In an effort to make the \ac{ME-IRL} comparison fair, we paid careful attention to collecting informative demonstrations. For each unknown feature, we recorded a mix of 20 demonstrations about the unknown feature only (with a focus on learning about it), the known feature only (to learn a reward weight on it), and both of them (to learn a reward weight combination on them). We chose diverse start and goal configurations to trace the demonstrations.

\begin{figure*}
\centering
\begin{subfigure}{.3\textwidth}
  \centering
  \includegraphics[width=\textwidth,left]{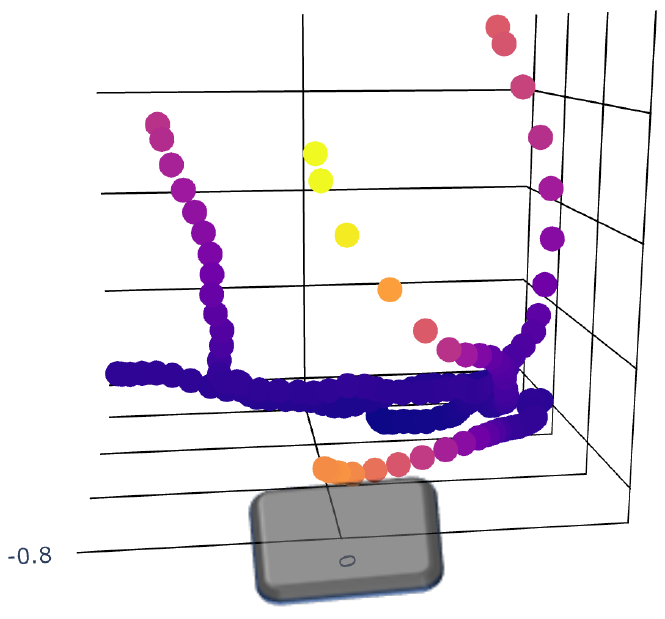}
\end{subfigure}%
\begin{subfigure}{.33\textwidth}
  \centering
  \includegraphics[width=\textwidth,left]{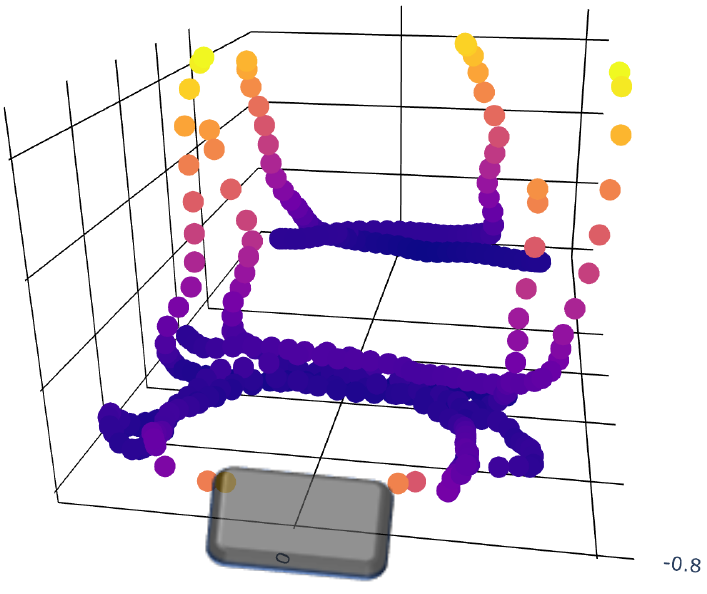}
\end{subfigure}
\begin{subfigure}{.33\textwidth}
  \centering
  \includegraphics[width=\textwidth,left]{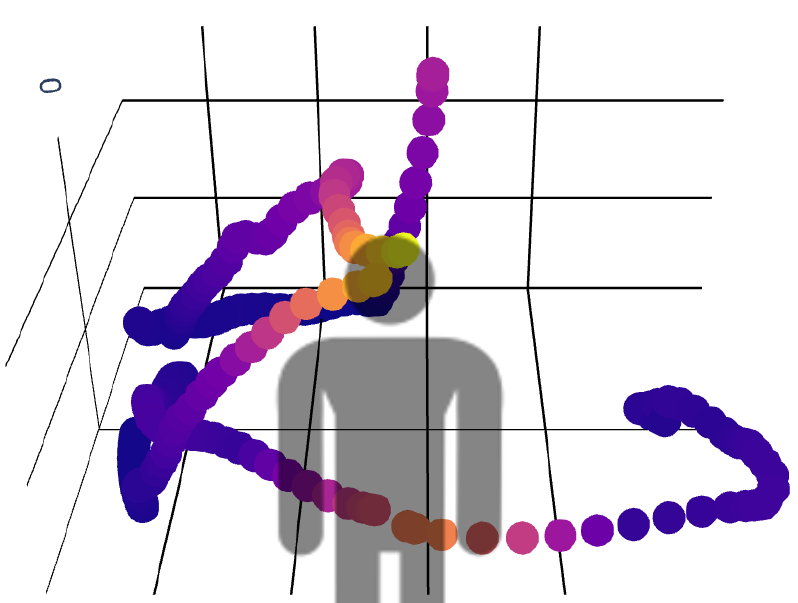}
\end{subfigure}
\caption{A few representative demonstrations collected for task 1 (left), task 2 (middle), and task 3 (right). The colors signify the true reward values in each task, where yellow is low and blue is high.}
\label{fig:Demos}
\end{figure*}

For task 1, we had a mix of demonstrations that start close to the table and focus on going around the laptop, ones that are far away enough from the laptop such that only staying close to the table matters, and ones where both features are considered. Fig. \ref{fig:Demos} (left) shows examples of such demonstrations: the two in the back start far away enough from the laptop but at a high height, and the two in the front start above the laptop at different heights.

For task 2, we collected a similar set of trajectories, although we had more demonstrations attempting to stay close to the table when the laptop was already far away. Fig. \ref{fig:Demos} (middle) shows a few examples: the two in the back start far away from the laptop and only focus on staying close to the table, a few more start at a high height but need to avoid the laptop to reach the goal, and another two start above the laptop and move away from it.

For task 3, the most difficult one, some demonstrations had to avoid the person slightly to their side, while others needed to avoid the person more aggressively in the front. We also varied the height and start-goal locations, to ensure that we learned about each feature separately, as well as together. Fig. \ref{fig:Demos} (right) shows a few of the collected demonstrations.

\subsection{Raw State Space Dimensionality}
\label{app:rawstate}

\begin{figure*}
\centering
\begin{subfigure}[b]{1\textwidth}
  \centering
  \includegraphics[width=0.9\linewidth]{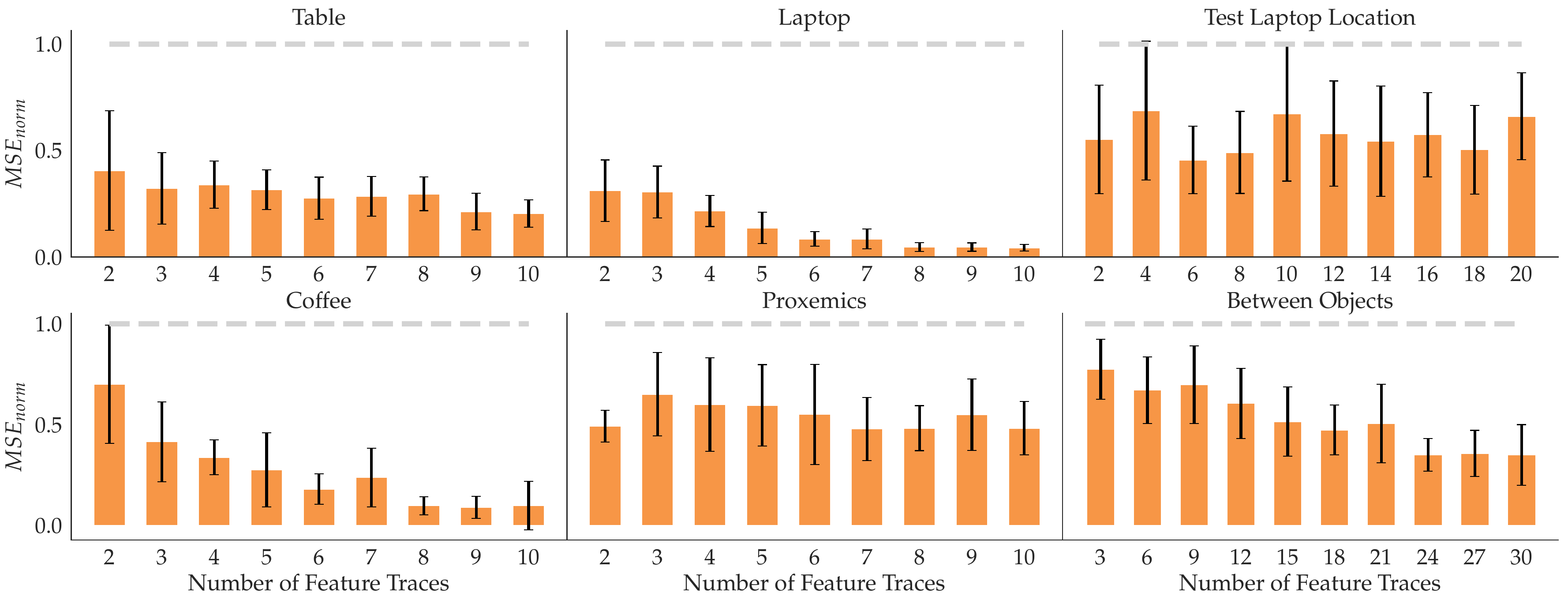}
\end{subfigure}%
\\
\begin{subfigure}[b]{1\textwidth}
  \centering
  \includegraphics[width=0.9\linewidth]{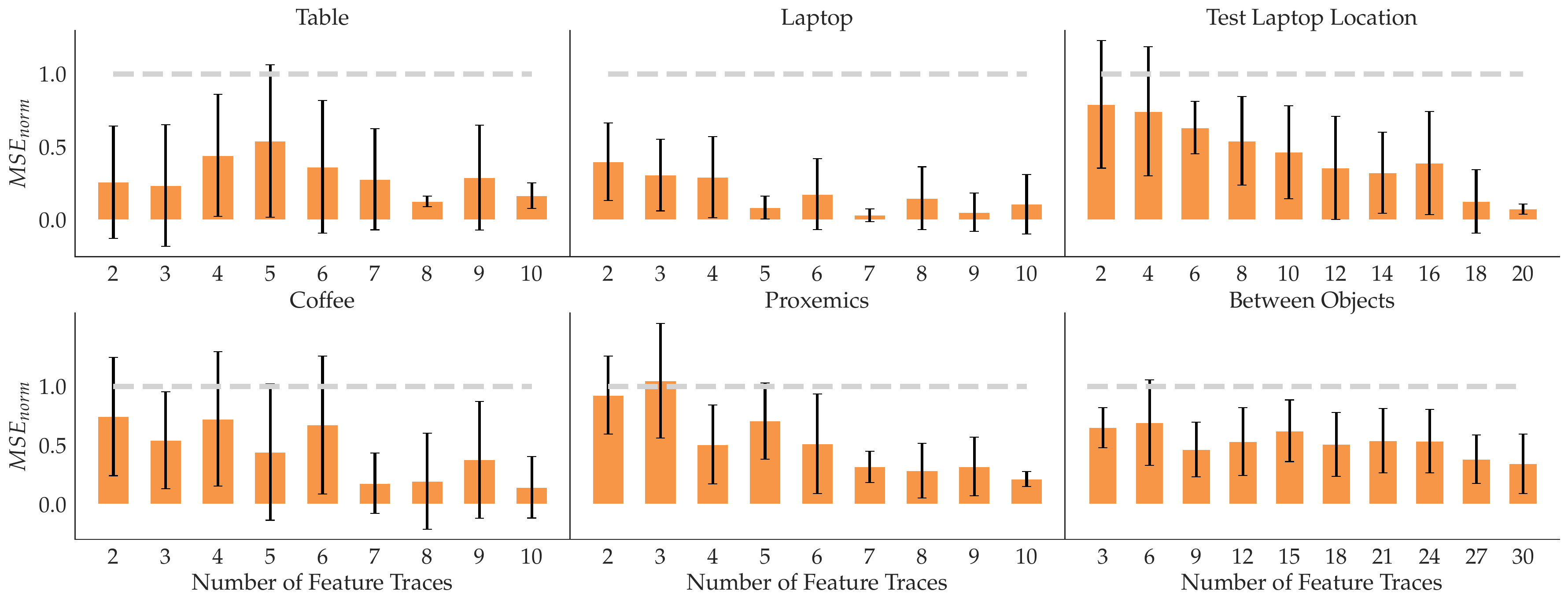}
\end{subfigure}
\caption{Quantitative feature learning results for 36D without (above) and with (below) the subspace selection heuristic. For each feature, we show the $\text{MSE}_{\text{norm}}$ mean and standard error across 10 random seeds with an increasing number of traces (orange) compared to random performance (gray).}
\label{fig:FERL_36_D}
\end{figure*}

In our experiments in Sec. \ref{sec:feature_expert} and \ref{sec:reward_experts}, we chose a 36D input space made out of 27 Euclidean coordinates ($xyz$ positions of all robot joints and environment objects) and 9 entries in the \ac{EE}'s rotation matrix. We now explain how we chose this raw state space, how spurious correlations across different dimensions can reduce feature learning quality, and how this adverse effect can be alleviated.

First, note that the robot's 7 joint angles and the $xyz$ positions of the objects are the most succinct representation of the state, because the positions and rotation matrices of the joints can be determined from the angles via forward kinematics. With enough data, the neural network should be able to implicitly learn forward kinematics and the feature function on top of it. However, we found that applying forward kinematics a-priori and giving the network access to the $xyz$ positions and rotation matrices for each joint improve both data efficiency and feature quality significantly. In its most comprehensive setting, thus, the raw state space can be 97D (7 angles, 21 $xyz$ coordinates of the joints, 6 $xyz$ coordinates of the objects, and 63 entries in rotation matrices of all joints).


Unfortunately, getting neural networks to generalize on such high dimensional input spaces, especially with the little data that we have access to, is very difficult. Due to the redundancy of the information in the 97D state space, the feature network $\phi_{\psi}$ frequently picks up on spurious correlations in the input space, which decreases the generalization performance of the learned feature. In principle, this issue could be resolved with more diverse and numerous data. Since we want feature learning to be as effortless as possible for the human, we instead opted for the reduced 36D state space, focusing directly on the $xyz$ positions and the \ac{EE} orientation.



Now, as noted in Sec. \ref{sec:feature_expert}, the spurious correlations in the 36D space still made it difficult to train on both the position and orientation subspaces. To better separate redundant information, we devised a heuristic to automatically select the appropriate subspace for a feature. For each subspace, the algorithm first trains a separate network for 10 epochs on half of the input traces and evaluates its generalization ability on the other half using the \ac{FERL} loss. The subspace model with the lower loss (better generalization) is then used for $\phi_{\psi}$ and trained on all traces. We found this heuristic to work fairly well, selecting the right subspace on average in about 85\% of experiments.

To test how well it works in feature learning, we replicated the experiment in Fig. \ref{fig:FERL_MSE} on the 36D state space, both with and without the subspace selection heuristic. A first obvious observation from this experiment is that performing feature learning on separate subspaces (Fig. \ref{fig:FERL_MSE}) results in lower MSEs for all features and $N$ number of traces than learning from all 36 raw states (Fig. \ref{fig:FERL_36_D}). Without the heuristic (Fig. \ref{fig:FERL_36_D} above), we notice that, while spurious correlations in the raw state space are not problematic for some features (\textit{table}, \textit{coffee}, \textit{laptop}, \textit{between objects}), they can reduce the quality of the learned feature significantly for \textit{proxemics} and \textit{test laptop location}. Adding our imperfect heuristic (Fig. \ref{fig:FERL_36_D} below) solves this issue, but increases the variance on each error bar: while our heuristic can improve learning when it successfully chooses the correct raw state subspace, feature learning worsens when it chooses the wrong one. 

In practice, when the subspace is not known, the robot could either use this heuristic or it could ask the human which subspace is relevant for teaching the desired feature. While this is a first step towards dealing with correlated input spaces, more work is needed to find more reliable solutions. A better alternative to our heuristic could be found in methods for causal learning, such as Invariant Risk Minimization~\cite{Arjovsky2019InvariantRM}. We defer such explorations to future work.

\section{Implementation Details}
We report details of our training procedures, as well as any hyperparameters used. We tried a few different settings but no extensive hyperparameter tuning was performed. Here we present the settings that worked best for each method. 
The code can be found at \href{https://github.com/andreea7b/FERL}{https://github.com/andreea7b/FERL}.

\subsection{FERL Training Details}
\label{app:FERL_implementation}

The feature function $\phi_{\psi}(\sysstate)$ is approximated by a 2 layer, 64 hidden units neural network. We used a leaky ReLu non-linearity for all but the output layer, for which we used the softplus non-linearity. We normalized the output of $\phi_{\psi}(\sysstate)$ every epoch by keeping track of the maximum and minimum output logit over the entire training data.
Following the description in Sec. \ref{sec:humaninput}, the full dataset consists of $|\mathcal{D}| = \sum_{i=1}^{N}\binom{(n^i+1)}{2} + 2 \binom{N}{2}$ tuples, where the first part is all tuples $(s,s',\{0, 1\})$ encoding monotonicity and the second part is all tuples $(s,s',0.5)$ encouraging indistinguishable feature values at the starts and ends of traces. Note that $\sum_{i=1}^{N}\binom{(n^i+1)}{2} >> 2 \binom{N}{2}$, hence in the dataset there are significantly fewer tuples of the latter than the former type. This imbalance can lead to the training converging to local optima where the start and end values of traces are significantly different across traces. We addressed this by using data augmentation (adding each $(s,s',0.5)$ tuple 5 times to $\mathcal{D}$) and weighing the loss from the $(s,s',0.5)$ tuples by a factor of 10. We optimized our final loss function using Adam for 100 epochs with a learning rate and weight decay of 0.001, and a batch-size of 32 tuples over all tuples.

\subsection{ME-IRL Training Details}
\label{app:MEIRL_implementation}

The reward $r_{\omega}(\sysstate)$ is approximated by a 2 layer, 128 hidden units neural network, with ReLu non-linearities. As described in Sec. \ref{sec:reward_experts}, we add the known features to the output of this network before linearly mapping them to $r_{\omega}(\sysstate)$ with a softplus non-linearity. While $D^*$ is given, at each iteration we have to generate a set of near optimal trajectories for the current reward $r_{\omega}(\sysstate)$. For that we take the start and goal pairs of the demonstrations and use TrajOpt \cite{schulman2013trajopt} to generate an optimal trajectory for each start-goal pair, hence $|D^*|=|D^\omega|$. At each of the 50 iterations we go through all start-goal pairs with one batch consisting of the $D^*$ and $D^\omega$ trajectories of one randomly selected start-goal pair from which we estimate the gradient as detailed in Sec. \ref{sec:reward_experts}. We optimize the loss with Adam using a learning rate and weight decay of 0.001.

\section{Additional Results}

\subsection{\textit{Between Objects} with 9D State Space Input}
\label{app:betweenobjects}

\begin{figure}[H]
\includegraphics[width=.45\textwidth]{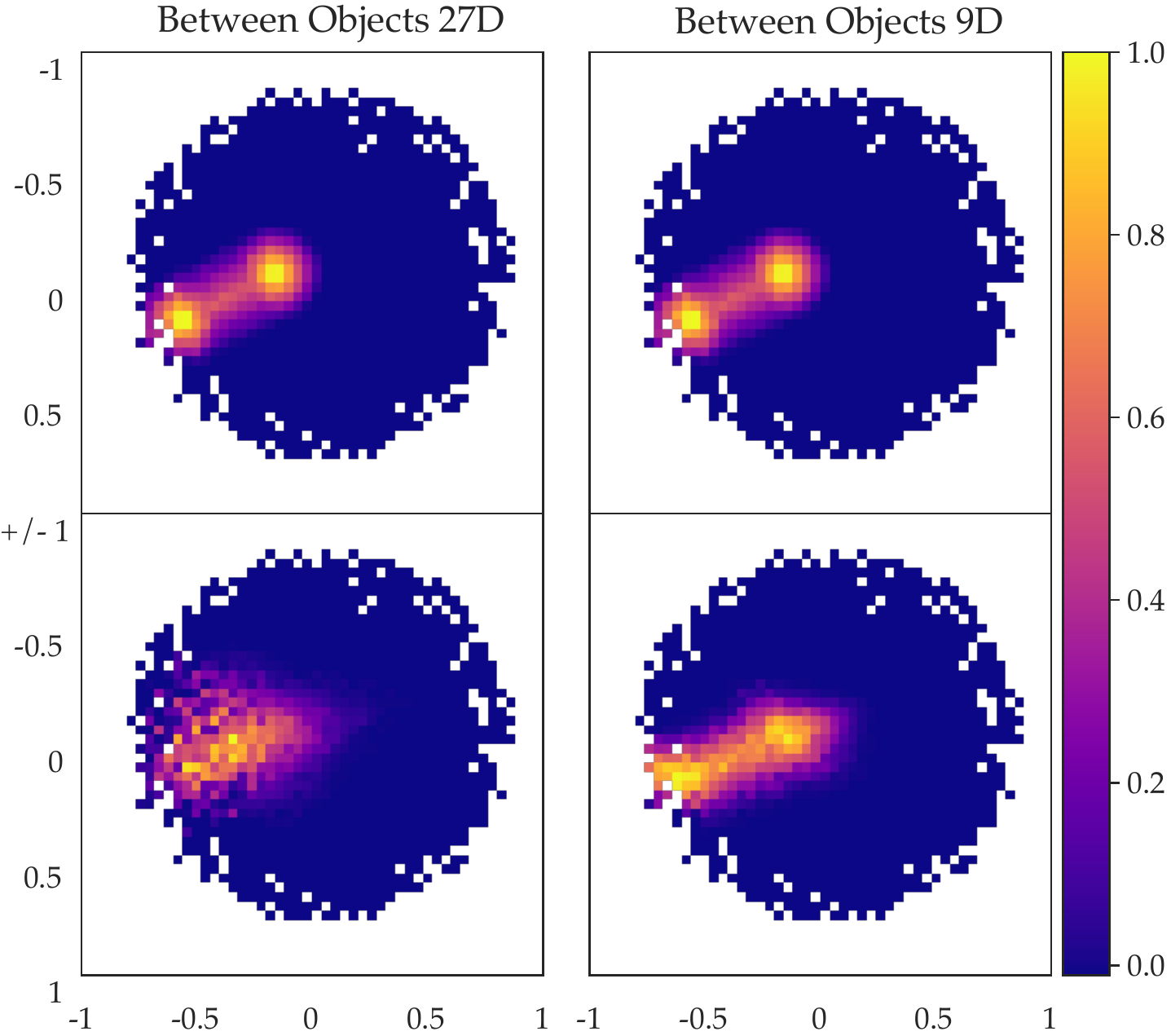}
\centering
\caption{The \textit{between objects} feature. (Left) Using a 27D highly correlated raw state space ($xyz$ positions of all robot joints and objects), the learned feature (Down) does not capture the fine-grained structure of the ground truth (Up). (Right) When using only 9D ($xyz$ positions of the \ac{EE} and objects), the quality of the learned feature improves.}
\label{fig:VisualComparisonC1}
\end{figure}

In Fig. \ref{fig:FERL_Qual} we saw that for \textit{between features}, while \ac{FERL} learned the approximate location of the objects to be avoided, it could not learn the more fine-grained structure of the ground truth feature. This could be an artefact of the spurious correlations in the high dimensional state space. To further analyze this result, we trained a network with only the dimensions necessary for learning this feature: the $xyz$ position of the \ac{EE} and the $xyz$ positions of the two objects. The result in Fig. \ref{fig:VisualComparisonC1} illustrates that, in fact, our method is capable of capturing the fine-grained structure of the ground truth; however, more dimensions in the state space induce more spurious correlations that decrease the quality of the features learned.

\begin{figure}[H]
\begin{subfigure}{.47\textwidth}
  \centering
  \includegraphics[width=\textwidth]{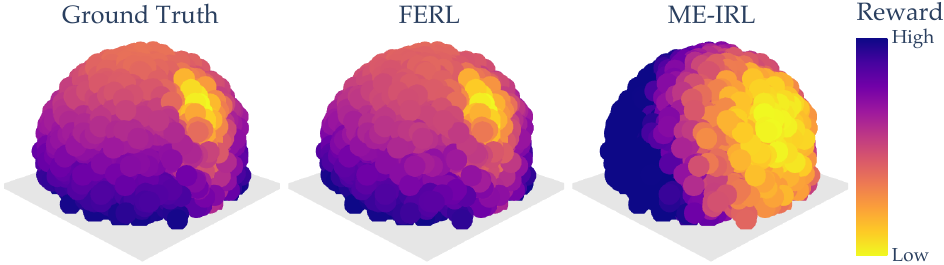}
\end{subfigure}%
\\
\begin{subfigure}{.47\textwidth}
  \centering
  \includegraphics[width=\textwidth]{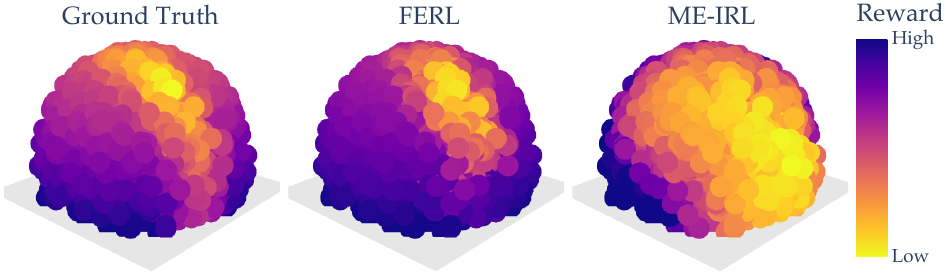}
\end{subfigure}
\caption{Visual comparison of the ground truth, \ac{FERL}, and \ac{ME-IRL} rewards for tasks 2 (above) and 3 (below).}
\label{fig:VisualComparisonC2}
\end{figure}

\subsection{Reward Visualization for Tasks 2 and 3}
\label{app:cases23}

Following the procedure detailed in Sec. \ref{sec:reward_experts}, we qualitatively compare the ground truth reward and the learned \ac{FERL} and \ac{ME-IRL} rewards. Figure \ref{fig:VisualComparisonC2} (above) shows the rewards for task 2, where the \textit{table} feature is unknown and $r_{\text{true}}=(0, 10, 10)(\phi_{\text{coffee}}, \phi_{\text{laptop}}, \phi_{\text{table}})^T$, and Figure \ref{fig:VisualComparisonC2} (below) for task 3 with the \textit{proxemics} feature unknown and $r_{\text{true}}=(0, 10, 10)(\phi_{\text{coffee}}, \phi_{\text{table}}, \phi_{\text{proxemics}})^T$. Similar to task 1 (Fig. \ref{fig:VisualComparison}), we observe that in task 2 \ac{FERL} is able to learn a fine-grained reward structure closely resembling \ac{GT}. In task 3, for the more difficult \textit{proxemics}, \ac{FERL} with just 10 features traces still recovers most of the reward structure. In both tasks, \ac{ME-IRL} only learns a coarse structure with a broad region of low reward which does not capture the intricate trade-off of the true reward function. 